\documentclass[journal]{IEEEtai}
\let\labelindent\relax
\usepackage{amsmath,amsfonts}
\usepackage{algorithmic}
\usepackage{enumitem}
\usepackage{algorithm}
\usepackage{array}
\usepackage[caption=false,font=normalsize,labelfont=sf,textfont=sf]{subfig}
\usepackage{multirow}
\usepackage{amsthm}
\newtheoremstyle{Remark} 
{3pt} 
{3pt} 
{} 
{} 
{\bfseries} 
{:} 
{.5em} 
{} 

\theoremstyle{Remark}
\newtheorem{remarknn}{Remark}

\usepackage{textcomp}
\usepackage{stfloats}
\usepackage{url}

\usepackage[colorlinks,urlcolor=blue,linkcolor=blue,citecolor=blue]{hyperref}
\usepackage{verbatim}
\usepackage{graphicx}
\usepackage{cite}
\usepackage[flushleft]{threeparttable}

\usepackage{xcolor}

\begin{document}

\title{Robotic Perception of Transparent Objects: \\ A Review}

\author{Jiaqi Jiang, Guanqun Cao, Jiankang Deng, Thanh-Toan Do and Shan Luo 
\thanks{Manuscript received: 31st March, 2023; revised: 6th August, 2023. This work was funded in part by the EPSRC ViTac project (EP/T033517/2), and in part by the King's College London and China Scholarship Council Award.}
\thanks{Jiaqi Jiang and Shan Luo are with Department of Engineering, King's College London, London WC2R 2LS, United Kingdom. E-mails: (jiaqi.1.jiang, shan.luo@kcl.ac.uk).}
\thanks{Guanqun Cao is with the smARTLab, Department of Computer Science, University of Liverpool, Liverpool L69 3BX, United Kingdom. E-mail: (g.cao@liverpool.ac.uk).}%
\thanks{Jiankang Deng is with Imperial College London, London SW7 2AZ, United Kingdom. E-mail: (j.deng16@imperial.ac.uk)}
\thanks{Thanh-Toan Do is with Department of Data Science and AI, Monash University, Clayton, VIC 3800, Australia. E-mail: (toan.do@monash.edu).}
\thanks{This paragraph will include the Associate Editor who handled your paper.}
}

\markboth{IEEE Transactions on Artificial Intelligence, Vol. 00, No. 0, Month 2023}%
{Jiang \MakeLowercase{\textit{et al.}}: Robotic Perception of Transparent Objects: A Review}


\maketitle

\begin{abstract}
Transparent object perception is a rapidly developing research problem in artificial intelligence. The ability to perceive transparent objects enables robots to achieve higher levels of autonomy, unlocking new applications in various industries such as healthcare, services and manufacturing.
Despite numerous datasets and perception methods being proposed in recent years, there is still a lack of in-depth understanding of these methods and the challenges in this field. 
To address this gap, this article provides a comprehensive survey of the platforms and recent advances for robotic perception of transparent objects. 
We highlight the main challenges and propose future directions of various transparent object perception tasks, i.e., segmentation, reconstruction, and pose estimation. We also discuss the limitations of existing datasets in diversity and complexity, and the benefits of employing multi-modal sensors, such as RGB-D cameras, thermal cameras, and polarised imaging, for transparent object perception. Furthermore, we identify perception challenges in complex and dynamic environments, as well as for objects with changeable geometries.
Finally, we provide an interactive online platform to navigate each reference: \url{https://sites.google.com/view/transperception}.
\end{abstract}

\begin{IEEEImpStatement}
Overall, this survey provides a valuable resource for researchers and practitioners in the field of robotic perception of transparent objects. By systematically reviewing the platforms and methods for transparent object perception, it promotes a deeper understanding of the current state-of-the-art, open research questions, and potential applications in various domains. Transparent object perception is a fundamental capability for robots to effectively interact with and manipulate transparent objects, with potential to revolutionise industries such as manufacturing, healthcare, and biotechnology by improving efficiency, accuracy, and safety.
The benefits of this paper are threefold. Firstly, it is the first comprehensive review of transparent object perception, providing a foundational knowledge base for further research and development. Secondly, the interactive online platform enhances accessibility and facilitates knowledge sharing, allowing readers to easily navigate and access information. Finally, the paper identifies challenges and open questions for transparent object perception, which can guide future research in this area and foster innovation in the field. 
\end{IEEEImpStatement}


\section{Introduction}
\begin{figure}[t]
   \includegraphics[width=\linewidth]{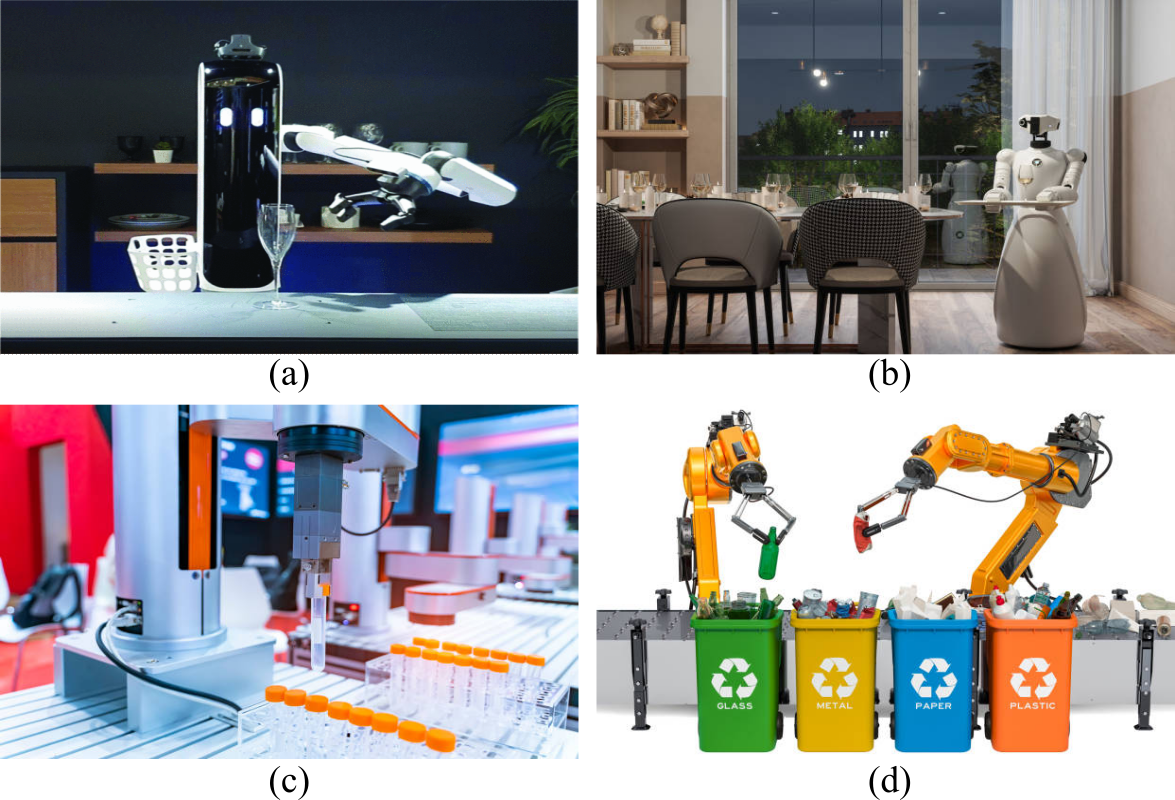}
  \caption{Typical applications of transparent object perception. (a) Robot assistant~\cite{chang2021ghostpose}(©[2021] IEEE); (b) Autonomous robot navigation; (c) Laboratory automation; (d) Waste sorting and recycling. }
\label{fig:applications}
\end{figure}

\IEEEPARstart{T}{ransparent} objects that transmit much of the light that falls on them and reflect little of it are ubiquitous in our daily lives such as glass windows and plastic bottles. They are also commonly used in various scenarios where robots are deployed, as shown in Fig.~\ref{fig:applications}. For instance, glass walls are prevalent in buildings for autonomous driving missions, and glass flasks are often used in automated research laboratories. Therefore, it is imperative for robots to accurately perceive, comprehend, and reason about these transparent objects in real-world environments.

However, perceiving transparent objects presents a significant challenge for robots due to their unique properties. Transparent objects lack salient surface features such as colour and texture, making their appearance highly dependent on the image background. Furthermore, the transparent materials of these objects violate the Lambertian assumption that optical 3D sensors (e.g., LiDAR and RGB-D cameras) are based on. The Lambertian assumption assumes that objects reflect light evenly in all directions, resulting in a uniform surface brightness from all viewing angles.
However, the surfaces of transparent objects both reflect and refract light, which breaks the Lambertian assumption. This property makes obtaining accurate depth data from depth sensors challenging, and the data is either invalid or contains unpredictable noise, further complicating the perception of transparent objects. 

To address the challenges of perceiving transparent objects, current research studies are focusing on two key problems: (1) locating transparent objects and (2) accurately estimating the depth of transparent objects. The first problem can be addressed in either 2D image space or the 3D real world, and is highly related to popular topics in the field of Computer Vision, such as transparent object segmentation~\cite{xie2020segmenting, Mei_2020_CVPR, mei2022glass}, and transparent object pose estimation~\cite{lysenkov2013pose, chang2021ghostpose, chen2022clearpose}.
The second problem is typically addressed by depth reconstruction methods that replace the noisy or invalid depth information of transparent objects with accurate and reliable depth estimates~\cite{sajjan2020clear, xu2021seeing, jiang2022a4t} or by 3D reconstruction methods that reconstruct the whole scene~\cite{ichnowski2021dex, kerr2022evo, dai2022graspnerf}. 
Thanks to the rapid development of Artificial Intelligence technology, recent years have seen tremendous progress in the development of new datasets for transparent objects and state-of-the-art approaches to the robotic perception of transparent objects. 
In fact, the number of publications in conferences and journals related to transparent object perception has significantly increased in the last three years.
Despite this progress, the most recent survey on transparent object perception~\cite{ihrke2010transparent} was published over a decade ago and primarily focused on reconstruction methods that were limited to controlled environments, rather than real-world robotic scenarios.  As such, there is a pressing need for more up-to-date surveys that can provide insights into the latest developments and advancements in transparent object perception for computer vision and robotics.

To advance the progress of the robotic perception of transparent objects, in this review article we comprehensively summarise the latest datasets and state-of-the-art perception methods for transparent objects, along with providing insightful remarks to facilitate the reader's understanding of the current status. 
We also provide insightful discussions of the remaining challenges and open questions. Furthermore, we offer an interactive online platform that allows users to explore various topics and methods presented in each reference included in this review paper, accessible at  \url{https://sites.google.com/view/transperception}. 
To the best of our knowledge, this is the first survey that specifically focuses on the robotic perception of transparent objects, aiming to provide a comprehensive and up-to-date resource for researchers and practitioners in this field.

\begin{figure}[t]
  \includegraphics[width=\linewidth]{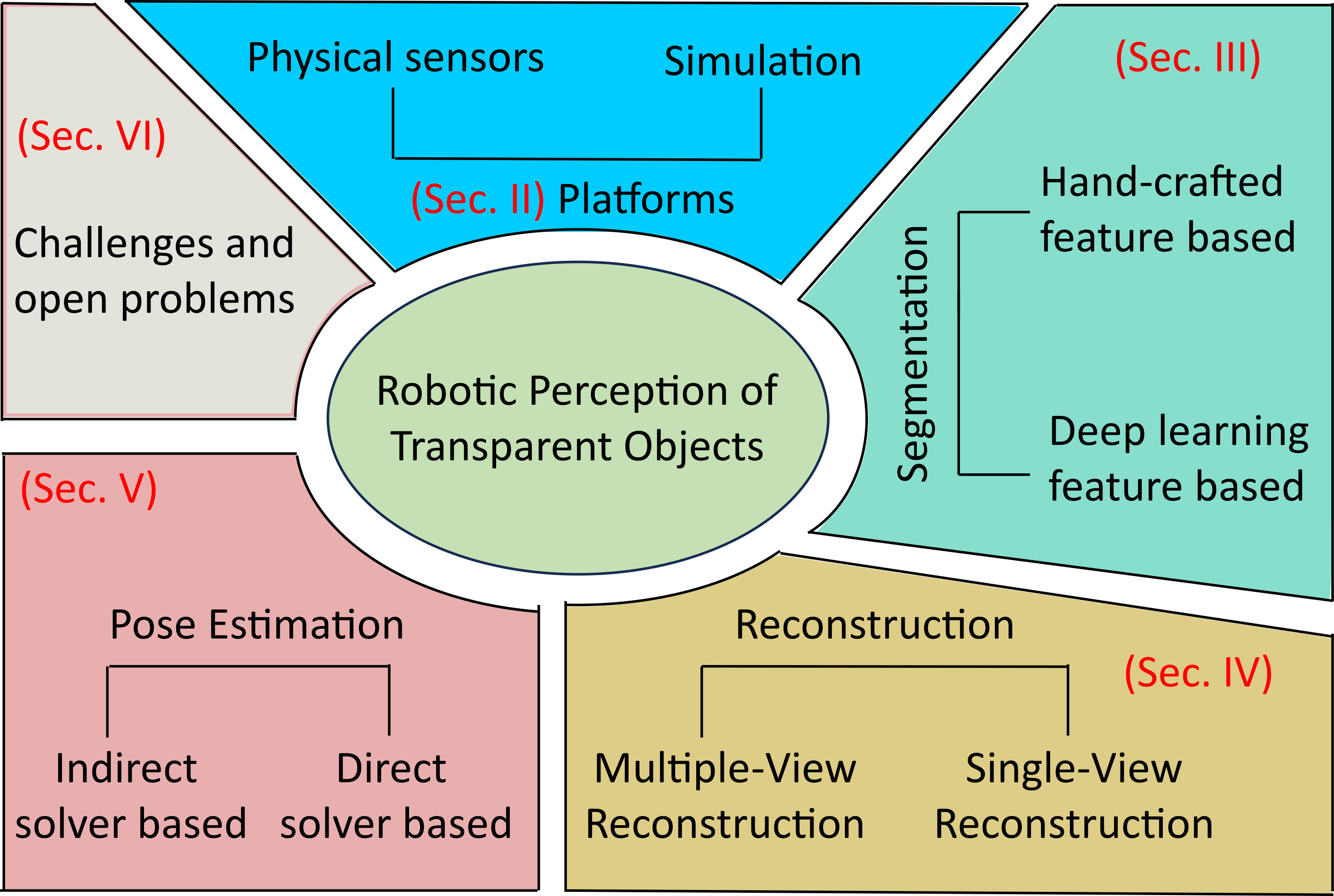}
  \caption{Schematic representation of the key components and methodologies discussed in this paper on robotic perception of transparent objects.}
\label{fig:structure}
\end{figure}

As is shown in Fig.~\ref{fig:structure}, the survey in this article is organised as follows: Section II introduces the platforms for transparent object perception from the perspectives of hardware and software. Then, in Sections III-V, we comprehensively review the current research status of three main tasks, i.e., segmentation, depth reconstruction and pose estimation. For each task-related section, we provide insightful discussions of the remaining challenges and open questions. Section VI summarises the challenges and open problems. 
Finally, Section VII concludes this article.
We independently review these three aforementioned robotic perception tasks for transparent objects since each task presents specific challenges that require distinct techniques and algorithms to overcome.

\section{Platforms} 
In this section, we will introduce an overview of the current platforms used for the robotic perception of transparent objects. We will start by introducing commercial sensors available in the market that have been widely adopted by researchers for transparent object perception. Next, we will discuss several popular simulation software and rendering engines used for generating synthetic datasets of transparent objects. By covering both commercial sensors and simulation-based approaches, this section aims to provide a comprehensive view of the tools and technologies available for transparent object perception.

\subsection{Sensors for Robotic Perception of Transparent Objects}
There are a variety of sensors used for transparent object perception, including RGB monocular cameras, RGB-D cameras, stereo cameras, light-field cameras, polarised cameras, RGB-Thermal cameras and tactile sensors, with some examples shown in Fig.~\ref{fig:sensors}.
While monocular RGB cameras are commonly used in object perception tasks, they are not suitable for transparent object perception since they only capture intensity information.
Thus, relying on these cameras alone for transparent object perception is inadequate, and alternative sensor types are necessary to complement or replace them in order to achieve robust and accurate results. For this reason, this article will focus on reviewing these alternative sensor types, which offer more comprehensive information (e.g., depth and temperature information) and are better suited for transparent object perception.

\subsubsection{RGB-D Cameras}

RGB-D cameras have proven to be a popular sensor choice for robotic perception tasks, such as object segmentation, 3D reconstruction and pose estimation. These cameras can capture both RGB images and per-pixel depth information, providing rich visual data for transparent object perception. Among the RGB-D camera options available, three main series have emerged as popular choices for researchers: the Kinect series from Microsoft, the Xtion series from Asus, and the RealSense series from Intel.

\begin{figure}[t]
  \includegraphics[width=\linewidth]{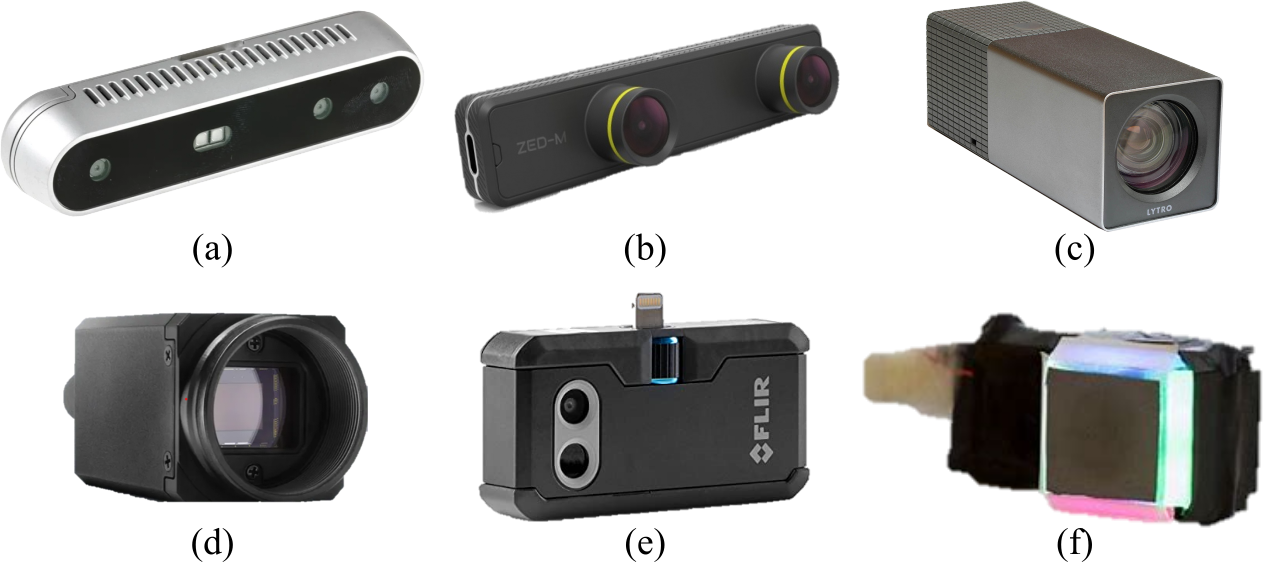}
  \caption{Example sensors used for transparent object perception. \textbf{(a)} RGB-D sensor provided by Intel~\cite{sajjan2020clear}; \textbf{(b)} ZED stereo camera provided by STEREOLABS~\cite{liu2020keypose}; \textbf{(c)} Light-field camera provided by Lytro~\cite{zhou2020lit}; \textbf{(d)} Polarised camera provided by LUCID~\cite{mei2022glass}; \textbf{(e)} RGB-Thermal Camera provided by FLIR~\cite{huo2022glass}; \textbf{(f)} GelSight tactile sensor used in~\cite{2022jiaqi}.}
\label{fig:sensors}
\end{figure}

\begin{figure}[t]
  \centering
   \includegraphics[width=0.8\linewidth]{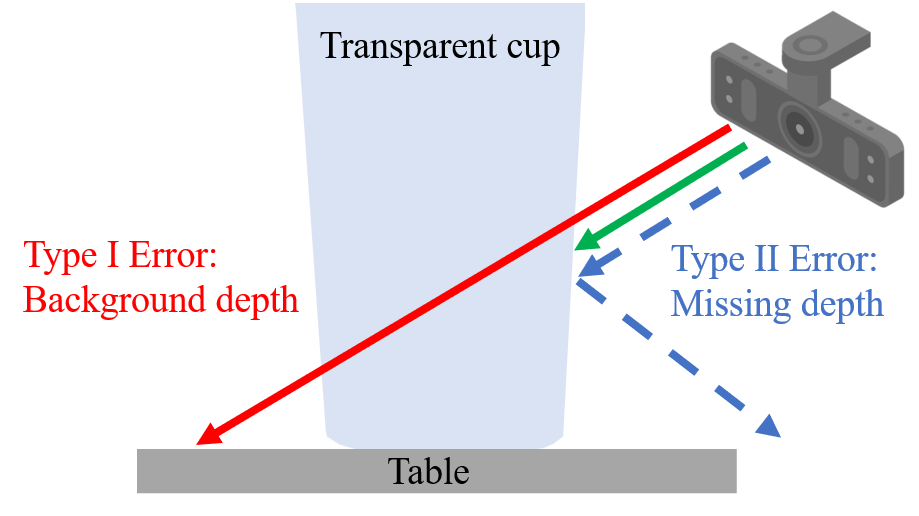}
  \caption{Two typical depth errors are shown when capturing the depth with commercial depth cameras. Type I errors (background depth estimates) result from capturing the background surfaces behind a transparent surface, as indicated by the red line; Type II errors (missing depth estimates) typically occur due to specular reflections on the transparent surface, represented by the blue dashed line. The green line denotes the accurate depth measurement.}
\label{fig:depth error}
\end{figure}

Taking the Intel RealSense D415\footnote{https://www.intelrealsense.com/depth-camera-d415/} in Fig.~\ref{fig:sensors}-(a) for example, it is based on the active infrared (IR) stereo principle, which uses an infrared laser projector to generate textures for the stereo cameras, leading to improved accuracy and more recent use for transparent object perception~\cite{sajjan2020clear, eppel2022predicting, 2022jiaqi}. However, transparent objects pose a challenge to depth estimation due to the violation of the Lambertian assumption, leading to two error types depicted in Fig.~\ref{fig:depth error}. Type I errors occur when light refracts through the transparent material and reflects back from the surface behind the object, leading to inaccurate depth estimates that correspond to the background depth (as shown in the case of the table in the figure). On the other hand, Type II errors arise due to specular highlights that alter the infrared patterns emitted by RealSense cameras. This change in patterns leads to incorrect stereo matching, resulting in missing depth information for the object.

\subsubsection{Stereo Cameras}
Stereo cameras are capable of simulating human bionic vision, as they utilise two or more lenses with a separate image sensor for each lens. This allows them to calculate depth information and enable 3D perception. For example, KeyPose~\cite{liu2020keypose} used the ZED\footnote{https://www.stereolabs.com/zed/} stereo camera shown in Fig.~\ref{fig:sensors}-(b) to predict the 3D positions of key points on transparent objects. In addition to ZED, there are several other stereo cameras available in the market, such as SceneScan 3D stereo vision sensors\footnote{https://nerian.com/products/scenescan-stereo-vision/} from Nerian vision technology, and Intel RealSense D405.

\subsubsection{Light-field Cameras} A light-field camera, also known as a plenoptic camera, is a specialised device capable of capturing comprehensive information about the light field emanating from a scene. Unlike traditional cameras that record only the light intensities at different wavelengths, light-field cameras can capture both the intensities and precise directions of light rays in space.
This additional data allows for a better understanding of the shape and position of transparent objects. 
The development of light-field cameras has been pioneered by Lytro, Inc., which has created a range of advanced devices, such as Lytro Light Field Digital Camera shown in Fig.~\ref{fig:sensors}-(c) and Lytro Illum. They have been used for transparent object recognition, segmentation, and pose estimation as demonstrated in \cite{maeno2013light,xu2019transcut2, zhou2020lit}. However, processing light-field data can be computationally intensive, which may limit their effectiveness and applicability.

\subsubsection{Polarised Cameras}
A polarised camera is a specialised type of camera that utilises a polarising filter to eliminate unwanted reflections and glare while enhancing contrast by colourising polarised angles of light. This unique feature enables polarised cameras to uncover hidden material properties, such as the reflective and refractive properties of transparent objects, providing superior visual clarity compared to standard RGB cameras. 

Polarisation data can be interpreted in two primary ways: the Degree of Linear Polarisation (DoLP) and the Angle of Linear Polarisation (AoLP). The DoLP measures the intensity of polarised light relative to the total intensity of light, while the AoLP indicates the orientation of the polarisation axis. By using polarised cameras, researchers and practitioners can obtain valuable insights into the properties of transparent materials, making them a valuable tool in transparent object perception as investigated in \cite{mei2022glass}.

Recently, Teledyne FLIR and LUCID have launched their Blackfly S polarised cameras\footnote{https://www.flir.com/products/blackfly-s-usb3}, and Triton\footnote{https://thinklucid.com/product/triton-5-mp-polarization-camera/} polarised cameras as shown in Fig.~\ref{fig:sensors}-(d), respectively.
Studies have shown that polarised cameras outperform conventional RGB cameras in transparent object segmentation tasks, as seen in recent works~\cite{kalra2020deep, mei2022glass}. 
However, due to their high prices (£2,000 for a Phoenix 5.0 MP Polarisation camera), they have yet to see widespread use in robotic applications.

\begin{figure}[t]
   \includegraphics[width=\linewidth]{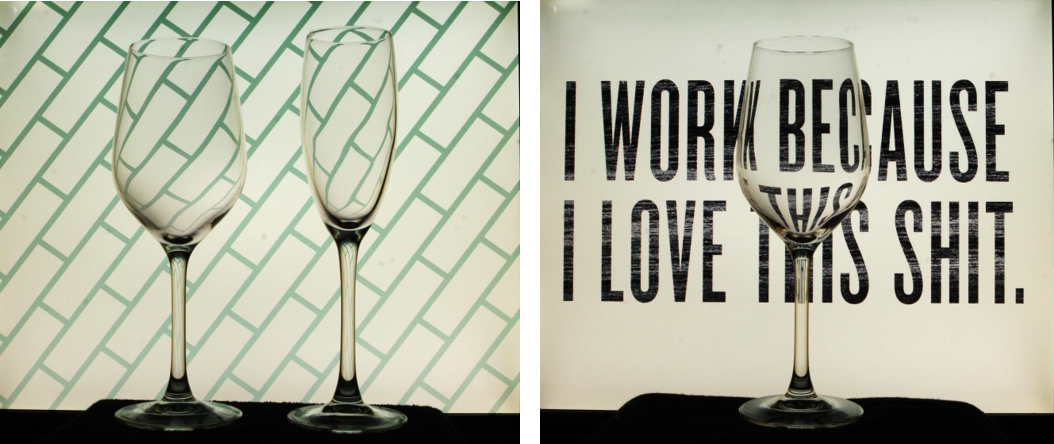}
  \caption{Diverse appearances of glass goblets under different backgrounds (collected from Tom-Net dataset~\cite{chen2018tom}). }
\label{fig:objects}
\end{figure}

\subsubsection{RGB-Thermal Cameras}
An RGB-Thermal camera combines an RGB camera module that can capture visible light with an infrared camera module that can detect thermal energy. Transparent materials can be opaque to thermal radiation in the range of 8 to 12 µm, due to absorption and scattering of thermal radiation by atomic bonds, which are not visible in the visible spectrum~\cite{larkin2017infrared}. Hence, compared to traditional RGB cameras, thermal cameras capture fewer arbitrary textures on transparent surfaces such as glass. This unique advantage can be helpful in dealing with the diverse appearance of transparent objects, as illustrated in Fig.~\ref{fig:objects}.

However, the price of thermal cameras can vary greatly depending on their type, resolution, and features. High-end thermal cameras, like the FLIR A65, can cost tens of thousands of dollars, while lower-end thermal cameras like the FLIR ONE Pro\footnote{https://www.flir.co.uk/products/flir-one-pro} shown in Fig.~\ref{fig:sensors}-(e) can be purchased for just a few hundred dollars. The resolution of thermal cameras also varies widely, ranging from as high as 640$\times$512 pixels to as low as 80$\times$60 pixels.

\subsubsection{Tactile Sensors}
A tactile sensor is a type of sensor that can detect and measure information physical interactions, such as pressure and force, without being affected by environmental lighting conditions~\cite{luo2017robotic}. As tactile sensors are not influenced by the diverse appearance of transparent objects, they can provide accurate data in any environment, making them a good complement to cameras for sensing challenging transparent objects. For example, Zhang et al.~\cite{zhang2022multimode} used a tactile sensor to classify the materials of transparent objects. The GelSight~\cite{yuan2017gelsight} tactile sensor shown in Fig.~\ref{fig:sensors}-(f) has been used in~\cite{2022jiaqi} to assist a camera for transparent object grasping. However, tactile sensors are usually small and designed to detect contact at a specific point or small area. As such, they are often used in conjunction with other remote-sensing devices, such as cameras, to provide a more comprehensive understanding of the environment. Furthermore, the cost of tactile sensors can range from a few pounds for a basic pressure-sensitive button to several thousand pounds for a high-resolution, multi-channel tactile sensor array~\cite{wettels2008biomimetic}.

\begin{remarknn}
In this subsection, we have reviewed and summarised six different sensors for transparent object perception, as listed in Table \ref{tab:sensors}. With the exception of tactile sensors, all of these sensors are capable of sensing a medium or large field, which makes them suitable for the remote perception of transparent objects. However, when the transparent objects are located at a significant distance, RGB-D cameras and light-field cameras may not be as effective as other sensors due to limitations in their sensing ranges.  

It is important to note that some sensors may struggle to capture or detect visual information in environments with low levels of ambient light, such as in dimly lit rooms or at night. Stereo cameras and light field cameras, for example, rely on visible light and may be less effective in such conditions. In contrast, sensors like RGB-Thermal cameras and polarised cameras can capture images of transparent objects in low-light conditions with good contrast and clarity, thanks to their distinct imaging principles. 

Furthermore, there exists a trade-off between the performance and cost of sensors. High-performance sensors, such as polarised sensors and RGB-Thermal cameras, often have the ability to sense a large range and function in low-light conditions, but their cost is typically high. On the other hand, low-cost sensors, such as RGB cameras, may not perform as well in challenging environments, but they are more accessible for researchers and developers. It is important to consider the specific requirements and constraints of a given application when selecting a sensor for transparent object perception.
\end{remarknn}

\begin{table}[t]
	\centering
        
 \begin{threeparttable}
		\caption{Comparison of different sensors used for transparent object perception.}
		\label{tab:sensors}
		\begin{tabular}{c | c | c  | c |c }
        \hline
        Sensor Type &  Feature & Range  & Night Vision  &Price  \\
        \hline
        \hline
        RGB-D &  Noisy depth  & M & Good & \pounds\pounds\\
        \hline
        Stereo  & Light disparity & L  & Poor &\pounds\pounds\\
        \hline
        Light-field   & Light distortion & M  & Poor &\pounds\pounds\\
        \hline 
        Polarised    & AoLP, DoLP & M-L   & Good &\pounds\pounds\pounds\\
        \hline
        RGB-T & Temperature  & L   & Good &\pounds\pounds-\pounds\pounds\pounds\\
        \hline 
        Tactile   & Contact shape & S & N/A &\pounds-\pounds\pounds\pounds\\
        \hline

  \end{tabular}
  
    \begin{tablenotes}
        \small
        \item Note: L, M, and S represent large, medium and small detection ranges, respectively. \pounds\pounds\pounds, \pounds\pounds, and \pounds~represent the price from highest to lowest.
        \end{tablenotes}
    \end{threeparttable}
  
\end{table}

\subsection{Platforms for Synthetic Dataset Generation} \label{2:2}
When obtaining and annotating real-world data is challenging and time-consuming, synthetic dataset generation provides an alternative solution. Advances in computer graphics and robotics have led to the development of simulation software in recent decades, such as Gazebo~\cite{koenig2004design}, V-REP~\cite{rohmer2013v}, Unreal Engine~\cite{qiu2016unrealcv}, Blender\footnote{https://www.blender.org/}, and Omniverse\footnote{https://www.nvidia.com/en-gb/omniverse/}. 

Many simulation software applications include an integrated rendering engine that calculates the interaction of light with the objects in the virtual scene to create a realistic image and is responsible for generating the final synthetic images or videos. Some software may also support additional external rendering engines that can be installed and used for more advanced or specialised rendering tasks. However, it should be noted that certain simulation software, such as Gazebo and V-REP, does not support the rendering of transparent objects.

By comparing the software in Table~\ref{tab:blender}, readers can easily select a rendering engine that aligns with their research requirements.
In general, there are a few key features that are desired for simulating transparent objects: 

\begin{itemize}
    \item Speed. The simulation software should be able to generate images or videos quickly to minimise the time required for dataset generation. This is especially important when simulating large scenes or generating datasets with many images.
    \item Quality of the output images. The simulation software should be able to generate high-quality images or videos that accurately represent the appearance of transparent objects. This requires the rendering engine to simulate the interactions of light with transparent objects in a physically accurate manner. 
    \item Capable of simulating the artefacts (e.g., caustics) that occur as light travels through transparent objects. 
    Caustics are the patterns of focused light that are formed when light rays pass through or reflect off a curved or refractive surface, and often seen as bright, concentrated spots or streaks of light on these surfaces. 
    \item Support of robotics physics. Since simulation software is often used to generate datasets for robotics applications, it should also support the simulation of physics. This includes simulating the motion of transparent objects, the behaviour of joints and motors, and other physical properties relevant to robotics.
\end{itemize}

In this paper, we will introduce and evaluate a selection of widely used simulation software and their integrated rendering engines that can achieve near-photorealistic rendering of transparent objects.

\subsubsection{Blender}
Blender is a popular computer graphics software for generating synthetic datasets due to its free and open-source nature. In recent years, it has been increasingly used in various robotic applications, such as segmentation~\cite{denninger2020blenderproc, eppel2022predicting}, reconstruction~\cite{sajjan2020clear, ichnowski2021dex}, and pose estimation~\cite{zhang2022transnet, chen2022clearpose}. 

Blender offers several different rendering engines for transparent object rendering, e.g., Eevee, Cycles, and LuxCoreRender\footnote{https://luxcorerender.org/}. Eevee provides the fastest rending speeds but often produces unrealistic results. While the default transparent material in Cycles renders reasonably well, it is unable to produce physically accurate caustics or dispersion without the usage of bidirectional rendering, resulting in transparent objects still having shadows similar to opaque objects. 
However, with the recent release of Blender 3.2, Cycles now supports selective rendering of caustics in the shadows of refractive objects, albeit with only up to 4 refractive caustic bounces.

\begin{figure}[t]
  \includegraphics[width=\linewidth]{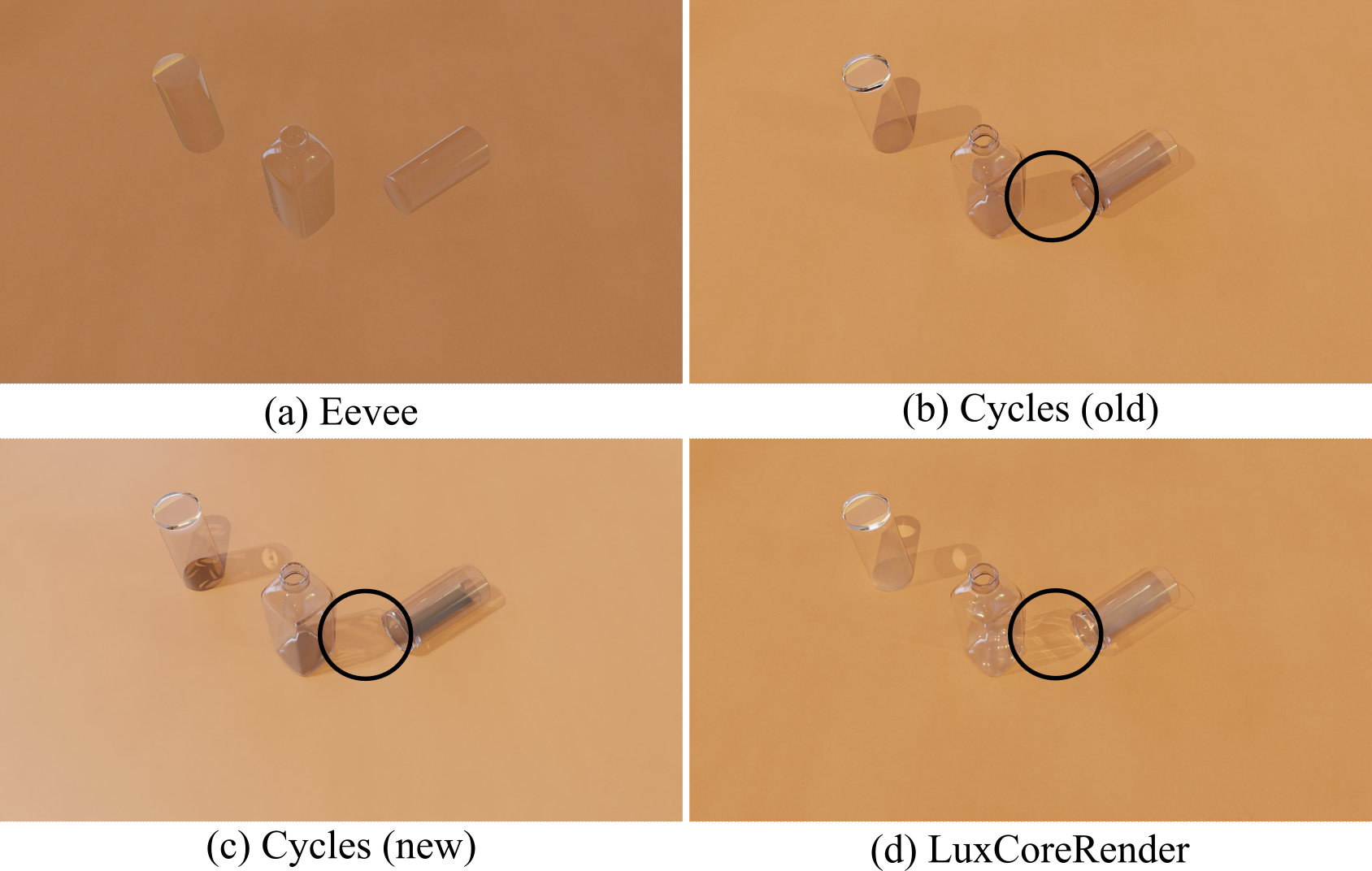}
  \caption{Comparison of various rendered images using different rendering engines in Blender. Both Eevee and Cycles (old) cannot accurately simulate complex optical effects such as caustics. The remaining two engines are capable of simulating caustics, but there are noticeable performance differences. Specifically, the LuxCoreRender engine outperforms Cycles (new) in terms of caustic rendering quality, as highlighted by the black circles. However, this comes at the cost of a higher computational expense.}
\label{fig:blender}
\end{figure}

In addition to the two built-in rendering engines, Blender users can also manually install other rendering engines. One such powerful engine is LuxCoreRender, which simulates the flow of light according to physical equations, producing photorealistic images of transparent objects. Fig.~\ref{fig:blender} compares the rendered images using different rendering engines in Blender. The figure clearly demonstrates that Eevee and
Cycles (old) are unable to simulate artefacts such as caustics, while Cycles (new) and LuxCoreRender can simulate caustics well.  

\begin{table}[t]
	\centering
		\caption{Comparison of different rendering engines used in simulating transparent objects.}
		\label{tab:blender}
        \scalebox{0.9}{
		\begin{tabular}{c | c | c | c | c }
        \hline
        Rendering Engine & Speed & Quality & Caustics & Robotic Physics \\
        \hline
        \hline
        Eevee & Fast & Low & $\times$ & $\times$\\
        \hline
        Cycles (old) & Slow & High & $\times$ & $\times$\\
        \hline
        Cycles (new) & Slow & High & $\surd$ & $\times$\\
        \hline 
        LuxCoreRender  & Slow & High& $\surd$ & $\times$\\
        \hline
        Unreal Engine & Fast & Medium & $\surd$ & $\surd$\\
        \hline 
        RTX Real-Time & Fast & Low & $\times$  & $\surd$\\
        \hline
        RTX Path-Traced & Medium & Medium & $\times$  & $\surd$\\
        \hline 
        RTX Iray &  Slow & High & $\surd$ & $\surd$ \\
        \hline 
        POV-Ray & Fast & Medium & $\surd$ & $\times$\\
        \hline

  \end{tabular}}
\end{table}

\subsubsection{Unreal Engine}
Unreal Engine (UE) is a 3D computer graphics game engine developed by Epic Games, which offers advanced Virtual Reality (VR), rendering, and physics capabilities. It has been widely used as the foundation for many robotic simulators, e.g., AirSim for autonomous unmanned vehicles~\cite{shah2018airsim}. Unlike Blender, which uses open-source rendering engines, UE implements its own rendering system directly with DirectX 11 and DirectX 12 pipelines. This includes a range of advanced features such as deferred shading, global illumination, lit translucency, and post-processing. In addition, UE includes GPU particle simulation that utilises vector fields, which can create stunning visual effects that add to the overall realism of the simulation.

\begin{figure}[t]
  \centering
  \includegraphics[width=\linewidth]{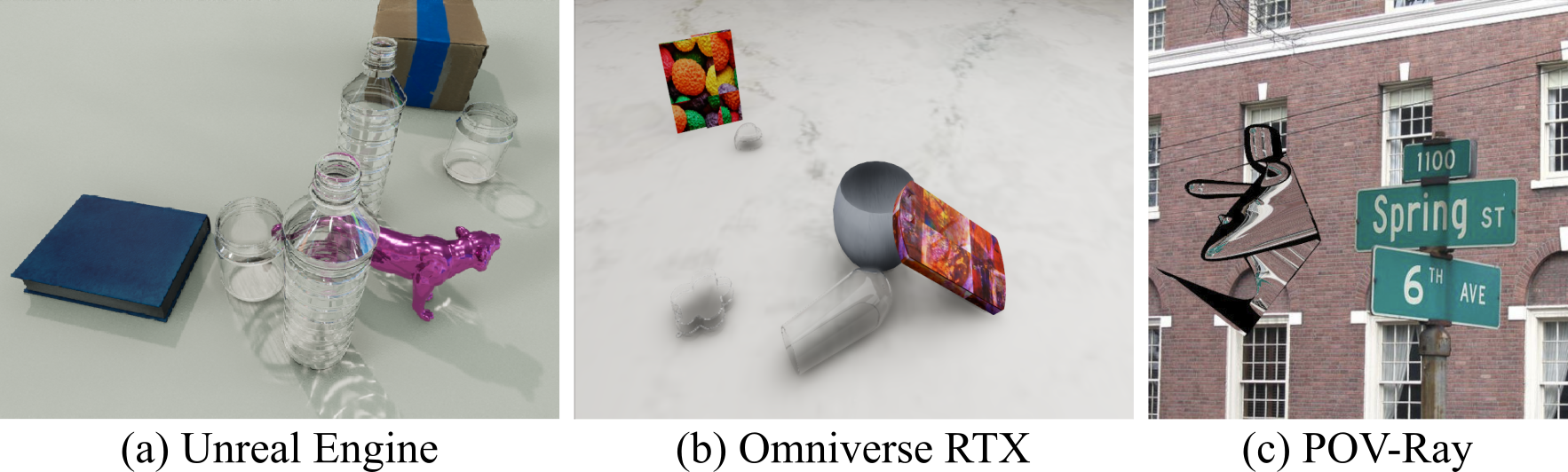}
  \caption{Comparison of synthetic images rendered with other render engines. \textbf{From left to right:} the examples are rendered with Unreal Engine, Omniverse RTX and POV-Ray, respectively.}
\label{fig:other render}
\end{figure}

\subsubsection{Omniverse}
Omniverse is a real-time graphics collaboration platform created by Nvidia. It has been used for applications in the visual effects and ``digital twin" industrial simulation industries.
Supported by NVIDIA's ecosystem which includes the NVIDIA PhysX engine, and NVIDIA Isaac Sim robotics simulation platform, Omniverse can easily render the scene with robotic arms or vehicles and achieve the physical interaction between robots and objects. Similar to Blender, Omniverse includes several different render engines: RTX Real-Time, RTX Path-Traced and RTX Iray which are used for creating the draft, previewing and finalisation, respectively.

\subsubsection{POV-Ray} The Persistence of Vision Raytracer (POV-Ray)\footnote{http://www.povray.org/} is a powerful and free software tool for creating high-quality 3D graphics. It allows users to create complex scenes, objects, and animations with great detail and realism. Unlike other 3D modelling software, which may focus more on ease of use and user interface, POV-Ray prioritises the quality of output images, and it achieves this by tracing the path of light rays through a virtual scene. This process of ray tracing simulates the physics of light as it interacts with objects in the scene, allowing for realistic reflections, refractions, and shadows. The source code is also available, making it an ideal choice for those who would like to modify and customise the software to their specific needs. 

\begin{remarknn}
In this subsection, a total of 4 simulation software with 8 integrated rendering engines are reviewed and summarised. Generally, Blender and Omniverse are the two most powerful software, both of which offer various rendering engines for different purposes, i.e., real-time rendering (Evee and RTX Real-Time) and photorealistic rendering (Cycles, LuxCoreRender and RTX Iray). 
However, there are two key differences between Blender and Omniverse: (1) Blender has attracted a larger and more active community of developers and users than Omniverse and is ideal for beginners, as it has been free and open-sourced for over two decades; (2) the recently developed Omniverse incorporates NVIDIA's PhysX and Isaac platforms, allowing for complex robotics simulations that involve handling transparent objects. Hence, for tasks like segmentation and depth reconstruction that do not necessitate robotic physics, LuxCoreRender is a strong recommendation due to its extensive documentation and ability to generate photorealistic images with the artefacts of transparent objects.

There also exists a trade-off between render quality and computational cost. While LuxCoreRender in Blender and RTX Iray in Omniverse are capable of producing high-quality images, they demand significant computational resources. When constrained by computing power, Unreal Engine and POV-Ray which have quick rendering speed and decent rendering quality can be alternative options. To assist the reader in understanding the rendering quality, multiple examples generated using UE, RTX Real-Time, and POV-Ray are provided in Fig.~\ref{fig:other render}. Moreover, several publicly available simulation environments specifically related to transparent objects such as the rendering code used in our previous work \cite{jiang2022a4t} and the SuperCaustics environment \cite{mousavi2021supercaustics} implemented in Unity could be found on our website\footnote{https://sites.google.com/view/transperception/platforms}.

\end{remarknn}

\section{Transparent Object Segmentation} \label{3}
Transparent object segmentation which categorises each pixel value of an image to be transparent or not is a crucial task in robotic perception. It allows autonomous robots to navigate in unknown environments such as a laboratory, market, or factory, without colliding with glass walls or windows. Furthermore, transparent object segmentation is a fundamental technique for other transparent object perception tasks, such as object pose estimation.

To achieve accurate transparent object segmentation, researchers have developed various methods that utilise machine learning techniques. In this section, we provide a comprehensive review of the most recent datasets published after 2015 for transparent object segmentation. We then summarise the state-of-the-art methods for transparent object segmentation. Finally, we highlight the main challenges of current transparent object segmentation methods from the perspective of both dataset generation and architecture designs.  

\subsection{Datasets}
In this subsection, we thoroughly summarise datasets published since 2015 for transparent object segmentation. We categorise the datasets based on the year, place of publication (Pub.), data type, number of objects in the images (\#Obj.), dataset size (\#Imgs), devices, scene type and object classes.

Overall, quite a few novel datasets were introduced in the past years, highlighting the need for advanced segmentation methods that can handle complex scenes and different lighting conditions. By providing a comprehensive summary of these datasets, we hope to facilitate the development of new and more accurate segmentation methods for transparent objects.

\begin{table*}
	\centering
        \begin{threeparttable} 
		\caption{Comparison of the datasets for transparent objects segmentation.}
		\label{tab:vision-dataset}
        
		\begin{tabular}{ | p{0.12\linewidth} | p{0.03\linewidth} | p{0.065\linewidth} | p{0.05\linewidth} | p{0.065\linewidth}| p{0.11\linewidth}| p{0.115\linewidth}| p{0.11\linewidth} |p{0.11\linewidth} |}
			\hline
			Dataset & Years & Pub. & \#Obj. & \#Imgs & Devices & Modality & Scene Type & Object Classes \\

        \hline
         TransCut~\cite{xu2015transcut} & 2015 & ICCV  & 7 & 49 (R) & Lytro camera & Light-field images & Single object & Glass containers \\
         \hline
         Tom-Net~\cite{chen2018tom} & 2018 & CVPR & 14 &178k (S), 876 (R) & POV-Ray, Digital camera & RGB images & Isolated objects & Glass containers, irregular glasses\\
         \hline
         GDD~\cite{Mei_2020_CVPR} & 2020 & CVPR & $\times$ & 3,916 (R) & Cameras, Smartphones & RGB images & Cluttered objects & Windows, glass walls and bulbs\\ 
         \hline
         Polarised~\cite{kalra2020deep} & 2020 & CVPR & 6 & 1,600 (R) & Stereo multipolar camera & RGB images, Polarised images & Cluttered objects & Plastic cups and trays, glasses\\
         \hline
         Trans10k~\cite{xie2020segmenting} & 2020 & ECCV & $\times$ & 10k (R) &  OpenImage, Smartphones & RGB images & Cluttered objects & Glass walls, plastic cups, etc.\\ 
         \hline
         GSD~\cite{lin2021rich} & 2021 & CVPR & $\times$ & 4,012 (R) & Other datasets, Phones & RGB images & Cluttered objects & Windows, glass walls\\
         \hline
         SuperCaustics~\cite{mousavi2021supercaustics}  & 2021 & ICMLA & 4 & 9k (S) & Unreal Engine & RGB images & Cluttered objects & Glass containers \\
         \hline
         TransProteus~\cite{eppel2022predicting}  & 2022  & Digital Discovery & 13k (S), 25 (R) & 50k (S), 104 (R) & Blender, RealSense D435 & RGB images, Depth images & Isolated objects & Glass containers with content\\ 
         \hline
         TransTouch~\cite{2022jiaqi} & 2022 &T-Mech & 9 & 9k (S), 180 (R) & Blender, RealSense D415 & RGB images & Isolated objects & Glass and plastic containers\\ 
         \hline
         RGBP-Glass~\cite{mei2022glass} & 2022 & CVPR & $\times$  & 4,511 (R) & LUCID PHX050S & RGB images, Polarised images & Cluttered objects & In-the-wild glass objects\\
         \hline
         RGB-T~\cite{huo2022glass} & 2023 & TIP & $\times$ & 5,551 (R) &  FLIR ONE Pro & RGB images, Thermal images & Cluttered objects & Windows, glass walls\\ 
         \hline
        \end{tabular}
        \begin{tablenotes}
        \small
        \item Note: $\times$ in \#Obj indicates that the dataset contains an unspecified number of objects, primarily consisting of similarly shaped glass walls and windows. S and R in \#Obj. and \#Imgs represent synthetic and real-world, respectively.
        \end{tablenotes}
        \end{threeparttable} 
\end{table*}

\noindent \textbf{TransCut}~\cite{xu2015transcut} dataset was collected in the real world using a light-field camera with 5$\times$5 viewpoints. This dataset includes seven transparent containers in different background scenes such as a library and a city backdrop. The objects are positioned about 50 cm from the camera, while the background images are positioned a further 100cm behind the objects.

\noindent \textbf{Tom-Net}~\cite{chen2018tom} dataset consists of 178k synthetic images generated with the POV-Ray rendering engine, and 876 real images captured using 14 transparent objects and 60 background images. To enhance the diversity of the dataset,  transparent objects were assigned a random refractive index $ \lambda \in[1.3, 1.5]$, and extensive data augmentation approaches such as colour augmentation and image scaling were carried out.

\noindent \textbf{GDD}~\cite{Mei_2020_CVPR} contains 3,916 pairs of glass and glass mask images, in which 2,827 images and 1,089 images are taken from indoor scenes and outdoor scenes, respectively. The images are captured with the latest cameras and smartphones, and the pixel-level glass masks are labelled by professional annotators. 

\noindent \textbf{Polarised}~\cite{kalra2020deep} uses a FLIR Blackfly S Monochrome Polar Camera to capture 1,600 images with 15 unique environments and 6 different transparent objects. To make the task more challenging, this dataset includes several 3D-printed objects which have the same shape as the transparent objects.  

\noindent \textbf{Trans10k}~\cite{xie2020segmenting} dataset contains 10,428 images that were either manually harvested from the internet (e.g., Google OpenImage) or captured with smartphones. The objects can be grouped into two categories of transparent objects, i.e., transparent things such as cups, bottles and glass; and transparent stuff such as windows, glass walls and glass doors. The dataset was relabelled finely in 2021, named Trans10k-v2.

\noindent \textbf{GSD}~\cite{lin2021rich} includes 4,012 real images with glass surfaces and corresponding masks. Similar to GDD~\cite{Mei_2020_CVPR}, the images are either from existing datasets and the Internet, or manually collected with a smartphone. As mentioned in~\cite{lin2021rich}, GSD includes objects with more complex shapes, which makes it more challenging.   

\noindent \textbf{SuperCaustic}~\cite{mousavi2021supercaustics} dataset is a synthetic transparent object dataset with 9,000 images generated in Unreal Engine.
Beyond the segmentation masks of caustics, it also provides estimated depths, surface normals, and object masks that can be used for other tasks such as depth reconstruction. Two main advantages of SuperCaustic are that it leverages the physics engine of Unreal Engine to arrange reasonable pose for each object, and it provides the control interface of caustic levels from soft to sharp. 

\noindent \textbf{TransProteus}~\cite{eppel2022predicting} dataset includes 50k synthetic images generated with Blender and 104 real-world images captured with a RealSense D435 camera. This synthetic dataset is one of the most challenging datasets for transparent object segmentation as it not only has high diversity (i.e., 13k different objects, 500 different environments) but also considers the challenging situations where simulated liquids are included.  

\noindent \textbf{TransTouch}~\cite{jiang2022robotic} dataset is a special dataset for segmenting the horizontal upper surface of transparent objects, which can be used for guiding a stable interaction between the robot and transparent objects. This dataset includes more than 9k synthetic images rendered with the LuxCoreRender engine and 180 real-world images captured with a RealSense D415 camera. 

\noindent \textbf{RGBP-Glass}~\cite{mei2022glass} is a polarisation glass segmentation dataset. Mei et al. use a trichromatic polariser-array camera (LUCID PHX050S) to collect a total of 4,511 RGB intensity images and corresponding pixel-aligned trichromatic images. It has been the most extensive publicly available RGBP-based dataset for glass-like object segmentation tasks.

\noindent \textbf{RGB-Thermal}~\cite{huo2022glass} (RGB-T) dataset is a glass segmentation dataset that consists of 5,551 pairs of RGB and thermal images, manually collected using a FLIR ONE Pro camera in a variety of scenes, such as libraries, shopping malls, and houses. While the raw thermal images have a resolution of 160$\times$120, they have been upscaled to 640$\times$480 using a super-resolution method. This dataset offers a unique challenge to segmentation algorithms, as it requires dealing with the differences between RGB and thermal imagery.

\begin{remarknn} In the above 11 public datasets for transparent object segmentation, most of them were collected using physical sensors in the real world, with the exception of a few synthetic datasets generated using simulators. We discuss the aforementioned datasets in terms of their size, the types of tasks, and their complexity.

\noindent \textbf{Dataset size.} 
Compared to real-world datasets (which range from 49 to 10k images), synthetic datasets are mostly larger in size (ranging from 9k to 100k images), but they lack the noise and inaccurate rendering models present in real-world data. Nevertheless, synthetic datasets are useful for Sim2Real transfer learning research. 

\noindent \textbf{Sensing modality.} While all of the datasets reviewed include RGB images, some of them offer additional sensing modalities. For example, RGBP-Glass~\cite{mei2022glass} not only includes RGB images but also trichromatic images collected by polarised cameras. The trichromatic images provide additional information about the polarisation of light passing through the glass. Similarly, RGB-Thermal~\cite{huo2022glass} includes thermal images, which capture temperature information and can be useful for identifying regions of glass that are either hotter or colder than their surroundings. By incorporating multiple sensing modalities, these datasets can offer more comprehensive information for transparent object segmentation, enabling researchers to explore new avenues for feature extraction and fusion.
\end{remarknn}

\begin{figure}[t]
  \includegraphics[width=\linewidth]{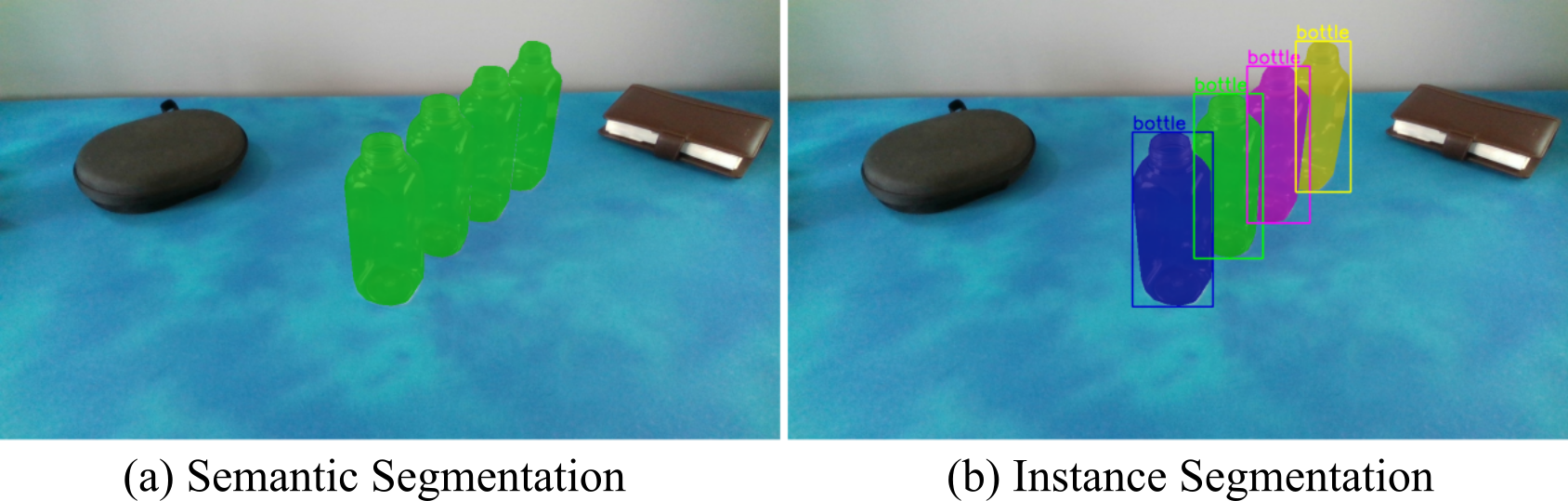}
  \caption{Transparent objects can be segmented in different ways. \textbf{(a)} Semantic segmentation~\cite{mei2022glass} of transparent bottles, where each pixel is assigned a semantic label, i.e., bottle in this case. \textbf{(b)} Instance segmentation~\cite{kalra2020deep}, where all pixels belonging to a single object are assigned a unique ID represented by a unique colour and different colours are used to represent each of the individual bottles.}
\label{fig:segmentation}
\end{figure}

\noindent \textbf{Task types.} As shown in Fig.~\ref{fig:segmentation}, there are two tasks in transparent object segmentation datasets: (1) semantic segmentation; (2) instance segmentation. Semantic segmentation that classifies pixels with semantic labels is essential for applications in autonomous driving. For example, autonomous robots need to avoid using the unreliable depth information of transparent objects for their self-localisation and avoid collisions with fragile transparent objects during navigation. By extending the scope of semantic segmentation, instance segmentation that partitions individual transparent objects is vital for other transparent object manipulation tasks. For example, the robots need to segment each object in order to be able to estimate their poses for manipulation. While all the datasets showcased in Table~\ref{tab:vision-dataset} are suitable for semantic segmentation of transparent objects, only Polarised~\cite{kalra2020deep}, SuperCaustics~\cite{mousavi2021supercaustics}, TransTouch~\cite{2022jiaqi}, and TransProteus~\cite{eppel2022predicting} datasets provide labels for instance segmentation of transparent objects.

\noindent \textbf{Dataset complexity.}
In terms of dataset complexity, a typical assessment method is by counting object classes. As TransProteus dataset~\cite{eppel2022predicting} uses 13k vessels with different contents for data collection, far exceeding other datasets, it can be recognised as the most challenging dataset. However, for datasets such as GDD~\cite{Mei_2020_CVPR}, Trans10k~\cite{xie2020segmenting}, and GSD~\cite{lin2021rich}, designed for the segmentation of glass walls and windows, the number of objects is difficult to define as most of them are plane surfaces with similar shapes. Hence, instead of comparing the number of objects in such datasets, we propose to use two evaluation metrics, i.e., Fractal Dimension (FD)~\cite{theiler1990estimating} and Mean Connected Components (MCC)~\cite{xie2020segmenting}, to quantitatively compare these datasets. Fractal dimension is a well-known measure for characterising geometric complexity and has been used in many object segmentation studies~\cite{ma2020rose}. Among these datasets compared in Table \ref{table:complexity}, GSD~\cite{lin2021rich} has the highest fractal dimensions and MCC, and can be recognised as the most challenging one.

\begin{table}[t]
	\centering
		\caption{Comparison of the dataset complexity.}
  \label{table:complexity}
    \begin{tabular}{  p{0.12\linewidth} | p{0.10\linewidth} | p{0.12\linewidth} | p{0.12\linewidth} | p{0.12\linewidth}| p{0.12\linewidth}}
    \hline 

    Metrics & GDD\cite{Mei_2020_CVPR} & Trans10k\cite{xie2020segmenting} & GSD\cite{lin2021rich} & RGBP-Glass\cite{mei2022glass} & RGB-T\cite{huo2022glass} \\
    \hline
    \hline
    FD~\cite{theiler1990estimating}& 0.887 & 1.043 & 1.076 & 0.925& 0.945\\ 
    \hline
    MCC & 1.95 & 3.96& 4.35 & 3.00 & 3.20\\
    \hline
    \end{tabular}

\end{table}

\subsection{Approaches for Transparent Object Segmentation}
In this subsection, we provide an overview of state-of-the-art transparent object segmentation methods developed over the past decade, focusing on both hand-crafted feature based and deep learning feature based approaches.

\subsubsection{Hand-crafted feature based} Considering that transparent objects without locally discriminative visual features and homogeneity of surface appearance, traditional local features~\cite{lowe1999object, bay2008speeded} are not applicable. Early studies~\cite{mchenry2005finding, mchenry2006geodesic} on transparent object segmentation mainly use visual cues such as the boundary features, and strong highlights in the surface to predict the regions of transparent objects. However, they only work well under the strong assumption that the background is similar on both sides of all glass edges~\cite{mchenry2005finding, mchenry2006geodesic}. 

To address the challenges posed by the transparency, some researchers consider introducing other modalities such as depth information~\cite{wang2012glass, wang2013glass, luo2015transparent}, and light-field images~\cite{xu2015transcut} to extract other hand-crafted features and further facilitate the RGB vision. In~\cite{wang2012glass, wang2013glass}, the distinctive patterns of missing depth that are caused by refraction and reflection were integrated into a Markov Random Field to segment the transparent objects. In~\cite{luo2015transparent}, the unknown depth areas were used to generate segmentation candidates with Grabcut~\cite{rother2004grabcut}. In~\cite{xu2015transcut}, light-field linearity was first used to find some initial candidate pixels of transparent objects, and then graph-cut optimisation was applied to obtain an accurate segmentation result.

\begin{table*}
	\centering
		\caption{Comparison of the deep learning methods for transparent object segmentation}
		\label{tab:segmentation}
        \scalebox{1}{
		\begin{tabular}{ | p{0.15\linewidth} | p{0.04\linewidth} | p{0.12\linewidth} | p{0.1\linewidth} | p{0.12\linewidth}| 
        p{0.10\linewidth}|
        p{0.07\linewidth}| p{0.07\linewidth}| }
			\hline
			Methods & Years  & Input Modality & Backbone & Model Structure &  Multiple Feature Fusion & Attention Module & Boundary Aware\\
            \hline
            Tom-Net\cite{chen2018tom} & 2018 & RGB &  VGG16  & U-Shaped & $\times$ & $\times$ & $\times$ \\
            \hline 
            Okazawa et al.\cite{okazawa2019simultaneous} & 2019 & RGB, Infrared & Unknown & DeepLab v3+ & $\checkmark$ & $\times$ & $\times$ \\
            \hline
            Mask R-CNN~\cite{madessa2019leveraging} & 2019 & RGB & ResNet-101 & R-CNN & $\checkmark$ & $\times$ & $\times$ \\
            \hline
            GDNet\cite{Mei_2020_CVPR} & 2020 & RGB  & ResNeXt-101 & Encoder-Decoder  & $\checkmark$  & $\checkmark$ & $\times$\\
            \hline      
            
            Polarised R-CNN~\cite{kalra2020deep} & 2020 & RGB, Polarisation & ResNet-101 &  R-CNN & $\checkmark$ & $\checkmark$ & $\times$  \\
            \hline
            TransLab\cite{xie2020segmenting} & 2020 & RGB & ResNet-50 & DeepLab v3+ & $\checkmark$ & $\checkmark$ & $\checkmark$ \\
            \hline 
            Lin et al.\cite{lin2021rich} & 2021 & RGB & ResNeXt-101 & Encoder-Decoder & $\checkmark$ & $\checkmark$ & $\checkmark$ \\
            \hline
            EBLNet\cite{he2021enhanced} & 2021 & RGB & ResNet-50 & DeepLab v3+ & $\checkmark$ & $\checkmark$ & $\checkmark$ \\
            \hline           
            Trans2Seg\cite{xie2021segmenting} & 2021 & RGB & ResNet-50 & Encoder-Decoder & $\times$ & $\checkmark$ & $\times$\\
            \hline
            FANet\cite{cao2021fakemix} & 2021 & RGB & ResNet-50 & U-Shaped & $\checkmark$ & $\checkmark$ & $\checkmark$\\
            \hline
            Xu et al.\cite{xu2021real} & 2021 & RGB & Darknet-53 & DeepLab v3+ & $\checkmark$ & $\times$ & $\times$ \\
            \hline
            P(rogress)GSNet\cite{yu2022progressive} & 2022 & RGB &ResNeXt-101 & U-Shaped & $\checkmark$&$\checkmark$ & $\times$  \\
            \hline
            Lin et al.\cite{linexploiting} & 2022 & RGB & SegFormer\cite{xie2021segformer} &  Encoder-Decoder&$\checkmark$& $\checkmark$&$\times$\\
            \hline
            Lin et al.\cite{lin2022depth} & 2022 & RGB, Depth & ResNeXt-101 &  Encoder-Decoder & $\checkmark$& $\checkmark$ & $\times$\\
            \hline
            Trans4Trans\cite{zhang2022trans4trans} & 2022 & RGB & Transformer & Encoder-Decoder & $\checkmark$ & $\checkmark$ & $\times$\\
            \hline
            P(olar)GSNet\cite{mei2022glass} & 2022 & RGB, Polarisation& Conformer\cite{peng2021conformer} & Encoder-Decoder&$\checkmark$& $\checkmark$&$\checkmark$ \\
            \hline
            Huo et al.\cite{huo2022glass}&2023&RGB, Thermal&ResNet-50& U-Shaped & $\checkmark$&$\checkmark$ & $\times$ \\
            \hline

        \end{tabular}}
\end{table*}

\subsubsection{Deep learning feature based}
Many researchers have focused on utilising deep neural networks for transparent object segmentation in the past years. In the following, we will discuss both single-modal and multi-modal methods that use deep learning features for transparent object segmentation. 

\noindent \textbf{Single-modal methods.} 
Single-modal methods employing deep learning techniques have been utilised to address transparent object segmentation challenges. These methods often leverage better latent feature extractors for glass segmentation. In~\cite{chen2018tom}, a U-shaped network that has the same spatial dimensions of features in the encoder layers and the decoder layers was used to segment transparent objects.   
In~\cite{madessa2019leveraging}, Mask R-CNN (regions with convolution neural networks) shown in Fig.~\ref{fig:segmentation}-(b), is used to detect each individual transparent object. Mei et al. in \cite{Mei_2020_CVPR} proposed a method named GDNet that utilised a large-field contextual feature integration module and a convolutional block attention module (CBAM) for feature fusion~\cite{woo2018cbam}. 
Xu et al. in~\cite{xu2021real} used dense connections between different atrous convolution blocks to restore more detailed information for glass segmentation. \cite{yu2022progressive} used multiple Discriminability Enhancement (DE) modules and Focus-and-Exploration Based Fusion (FEBF) to progressively aggregate features from high-level to low-level, implementing a coarse-to-fine glass segmentation.

Beyond applying different multi-level feature fusion modules, the design of the model structure is also researched. For example, Xie et al. in~\cite{xie2021segmenting} proposed a novel transformer-based segmentation pipeline named Trans2Seg to improve the learnt features. 
Zhang et al. in~\cite{zhang2022trans4trans} proposed a transparency perception model based on a dual-head Transformer named Trans4Trans, which has been integrated into a wearable assistive system for assisting the navigation of visually impaired people.
\cite{narasimhan2022self} achieved the segmentation of transparent liquid without requiring any manual annotations by using a generative model to translate coloured liquids to transparent liquids.

There are also several studies leveraging the boundary information for transparent object segmentation, as summarised in the last column of Table~\ref{tab:segmentation}. 
In~\cite{xie2020segmenting}, a boundary-aware segmentation method named TransLab was proposed that exploits boundaries as clues to improve the segmentation performance of vanilla DeepLabv3+ shown in Fig.~\ref{fig:segmentation}-(d). Similarly in \cite{cao2021fakemix}, a boundary-aware segmentation method was proposed with an adaptiveASPP module that captures features of multiple receptive fields. However, predicting the edge of objects with edge supervision in~\cite{xie2020segmenting, cao2021fakemix} may limit the generality of learning objects with various shapes. To enhance the boundary prediction, He et al.~\cite{he2021enhanced} proposed a network named EBLNet that utilised an edge-aware point-based graph convolution network module.
Lin et al. in~\cite{lin2021rich} utilised a Rich Context Aggregation Module (RCAM) to extract multi-scale boundary features and a reflection-based refinement module to differentiate glass regions from non-glass regions. Recently, Lin et al. in~\cite{linexploiting} proposed a method for addressing the problem of glass surface detection by integrating contextual relationships of scenes with spatial information, which is different from the other works that focus on low-level feature extraction, such as boundary and reflections.

\noindent \textbf{Multi-modal methods.} 
Instead of using only RGB images, there are a few publications leveraging multiple modalities for deep learning based transparent object segmentation.
In~\cite{okazawa2019simultaneous}, a three-stream encoder-decoder model was proposed that uses RGB images, infrared images, and concatenated RGB-IR images as input to recognise both transparent objects and semantic segmentation simultaneously. In~\cite{kalra2020deep}, a Polarised Mask R-CNN was proposed that uses three separate backbones and an attention fusion module to extract and merge the multi-modal features, i.e., vision and polarisation, respectively. 
Lin et al. in~\cite{lin2022depth} used a Cross-modal Context Mining (CCM) module to adaptively learn individual and mutual context features from RGB and depth information.  Mei et al.~\cite{mei2022glass} proposed PGSNet that dynamically fused both the trichromatic colour and polarisation cues. In~\cite{huo2022glass}, a neural network architecture was proposed that effectively combines an RGB-thermal image pair with a new multi-modal fusion module based on attention and integrates CNN and transformer to extract local features and long-range dependencies, respectively. 

\begin{remarknn}
In this subsection, we conduct a comprehensive comparison of various transparent object segmentation methods that use either hand-crafted features or deep learning based features. A notable trend shown in Table~\ref{tab:segmentation} is that early work primarily relied on RGB data, while recent studies increasingly utilise multi-modal information such as RGB images, polarised images, and thermal images. Additionally, feature extractors have evolved from the series of ResNet~\cite{he2016deep} to the series of vision Transformer~\cite{xie2021segformer, peng2021conformer}.

Moreover, as summarised in the last three columns of Table~\ref{tab:segmentation}, nearly all prior works utilise feature fusion and attention mechanisms to enhance their segmentation performance, while the use of complementary information like boundaries remains less explored. Hence, how to further exploit the complementary information and even the broader contextual information for transparent object segmentation could be a research problem to be further investigated.
\end{remarknn}

\begin{figure*}[t]
  \includegraphics[width=\linewidth]{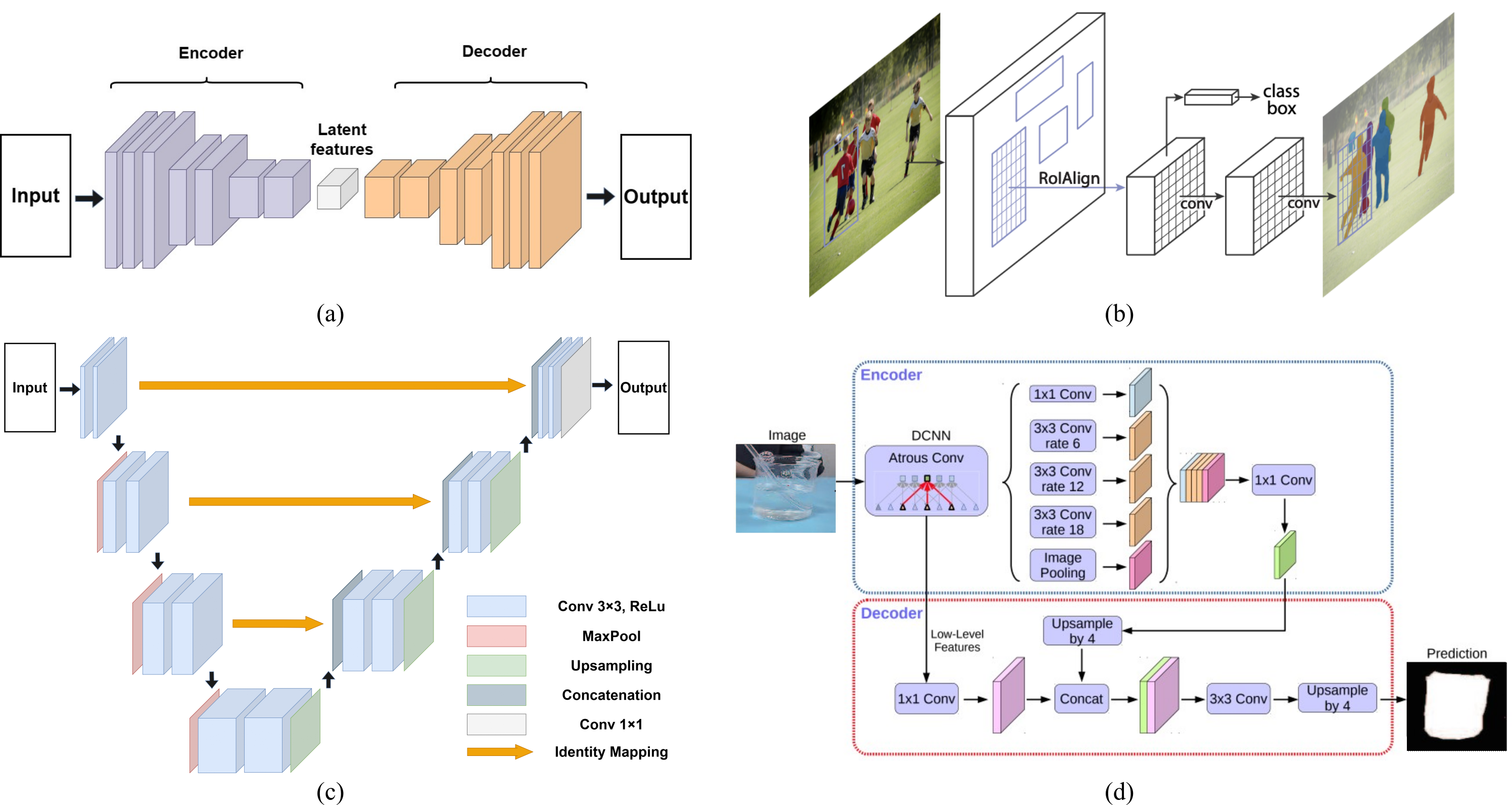}
  \caption{ \textbf{(a)} Encoder-Decoder Network\cite{goodfellow2016deep}; \textbf{(b)} Mask R-CNN\cite{he2017mask} ©[2017] IEEE; \textbf{(c)} U-Shaped Network~\cite{siddique2021u} ©[2021] IEEE; \textbf{(d)} DeepLabv3+\cite{chen2018encoder}.}
\label{fig:network}
\end{figure*}

\subsection{Challenges and Outlook}
The main challenges for transparent object segmentation are the limited scales of multi-modal datasets and the costly computation of large-scale segmentation architectures for transparent object segmentation. 

\subsubsection{Limited scales of transparent object datasets}
Training a deep neural network on a complex task requires a massive amount of data to overcome the over-fitting issue. Therefore, a set of large-scale datasets such as ImageNet~\cite{deng2009imagenet} and COCO~\cite{lin2014microsoft} were constructed via manual labelling for the recognition of common stuff in everyday scenes. However, transparent objects without salient boundaries significantly increase the labour intensity of pixel-wise annotation. For example, the most recent datasets in~\cite{mei2022glass} and~\cite{huo2022glass} only include 4k RGB-Polarisation images and 5k RGB-Thermal images, respectively. As a result, it is questionable how a transparent object segmentation model trained with a limited dataset performs under an unstructured or unseen environment.

\noindent \textbf{Data augmentation in simulation.}
Thanks to the rapid development of simulators for photo-realistic rendering, a set of studies~\cite{sajjan2020clear,nanbo2020learning,jiang2022robotic,eppel2022predicting} utilised simulators to render synthetic images with automatically generated annotations for transparent object segmentation. In this way, it not only can reduce the labour intensity of pixel-wise annotation but also can avoid the mistakes of manual labelling. 
However, synthetic image generation is not good enough to completely replace real-world images and suffers from several constraints. First, current state-of-the-art simulators such as Blender could only provide photo-realistic RGB-D image rendering, but cannot simulate other popular modalities that have been investigated in transparent object segmentation, such as polarisation images and thermal images. Second, the scene in the simulator is usually confined to the constructed indoor environment, as the construction of an outdoor environment such as commercial streets and big markets is very costly and time-consuming. Finally, the rendered images cannot be the same as the real-world images captured with physical cameras, which results in the Sim2Real domain gap. To this end, two promising directions to pursue in future research could be: (1) developing more functional simulators that can generate multiple modalities of data under various scenes; (2) learning domain-invariant features for Sim2Real transparent object segmentation.

\noindent \textbf{Efficient labelling tools.} Another way to overcome the limitations of transparent object datasets is by increasing the efficiency of data labelling. Tools such as LabelMe~\cite{russell2008labelme} required users to manually select the contours of the objects to be annotated, which results in a tedious and time-consuming labelling process. This manual process was improved by model-assisted approaches such as Deep Extreme Cut~\cite{maninis2018deep} and SuperPixel~\cite{achanta2012slic} which decreases the amount of user effort necessary to label images. Some other collaborative annotation tools either refine the partial annotations labelled by human annotators such as bounding boxes, and partial point clouds to fine annotations with pre-trained networks~\cite{lee2018leveraging}, or require human annotators to refine the rough annotations generated with pre-trained networks. However, it is still questionable whether the above semi-automatic methods could be applicable to transparent objects, as transparent objects do not have salient features and their appearance is inherited from the backgrounds. How to efficiently label data for transparent object segmentation is an important and challenging future direction. 

\noindent \textbf{Novel learning strategies.} Apart from the aforementioned solutions about data generation, it would be also interesting to leverage different learning strategies, such as using few-shot learning~\cite{dong2018few, liu2020crnet, zhang2021self} to train a robust model with a small amount of training data and using self-supervised learning~\cite{shimoda2019self, ziegler2022self} to train a model with partially or weakly labelled data.

\subsubsection{Transparent object segmentation in challenging environments}
Current state-of-the-art methods for transparent object segmentation show promising performance on datasets collected under good conditions. However, these conditions may not accurately reflect the complexities of real-world applications, such as autonomous driving, where dynamic environments and extreme lighting conditions are common. Therefore, it is crucial for these techniques to be robust to dynamic environments with moving transparent objects and a variety of lighting conditions ranging from very bright to very dim.

In dynamic environments, the main challenge of transparent object segmentation lies in facilitating real-time processing. There are several aspects we could look at for reducing the computational cost while not influencing the performance too much. First, it is more effective to utilise lightweight backbones for extracting features from sensing modalities that represent the physical properties of transparent objects such as depth, polarisation, and thermal, rather than employing ultra-deep networks for RGB image feature extraction. Additionally, future investigations could focus on accelerating multi-level or multi-modal fusion modules using techniques like pruning~\cite{liu2018rethinking} and quantisation~\cite{jin2020adabits}. Finally, segmenting transparent objects in a video that focuses on the use of temporal information between different frames could be also considered an important future direction.

Extreme lighting conditions pose significant challenges to transparent object segmentation. For instance, under low light, transparent objects may blend into the background due to low contrast. Conversely, high-intensity light can create misleading bright spots on them. Addressing these issues could involve training a robust multi-expert learning model across different lighting conditions, or utilising light-insensitive modalities like tactile sensing and polarised sensing.

\subsubsection{Network structures} 
Traditional segmentation networks like Mask R-CNN~\cite{he2017mask} and DeepLabv3+~\cite{chen2018encoder} shown in Fig.~\ref{fig:network} have become less popular among recent studies in transparent object segmentation. Instead, researchers are opting for newer encoder-decoder frameworks~\cite{lin2021rich, zhang2022trans4trans, mei2022glass} to enhance performance in transparent object segmentation. It is worth exploring whether current network structures can be replaced or augmented with generative models such as Generative Adversarial Networks (GANs)~\cite{goodfellow2016deep} or diffusion models~\cite{baranchuklabel}. These alternative models might bring new perspectives and techniques to the field, potentially leading to better segmentation outcomes. Additionally, it could be valuable to investigate the development of specialised models tailored for transparent objects.

\begin{table*}
	\centering
		\caption{Comparison of the datasets for transparent objects reconstruction}
		\label{tab:recon-dataset}
        \scalebox{1}{
		\begin{tabular}{ | p{0.12\linewidth} | p{0.03\linewidth} | p{0.05\linewidth} | p{0.04\linewidth} | p{0.08\linewidth}| p{0.12\linewidth}| p{0.11\linewidth}| p{0.22\linewidth} |}
			\hline
			Dataset & Years & Pub. & \#Obj. & \#Imgs  & Devices & Auto-collection & Special Feature \\
            \hline
            ClearGrasp\cite{sajjan2020clear} & 2020 & ICRA & 10 & 50k (S), 286 (R) & Blender, RealSense cameras  & $\checkmark$ (S), $\times$ (R) & Realistic synthetic image 
            \\
            \hline
            OOD\cite{zhu2021rgb} & 2021 & CVPR & 9 & 60k (S)  & Omniverse & $\checkmark$ (S) & Fast rendering
            \\
            \hline
            TODD\cite{xu2021seeing} & 2021 & CoRL & 6 & 15k (R)  & RealSense D415  & $\checkmark$ (R) & Apriltag arrays for auto collection 
            \\
            \hline
            Dex-NeRF\cite{ichnowski2021dex} & 2021 & CoRL & 5 (S), 6 (R) & 8 (S), 8 (R) & Blender, Cannon EOS, RealSense & $\checkmark$ (S), $\checkmark$ (R) & Challenging transparent objects
            \\
            \hline
            TransCG\cite{fang2022transcg} & 2022 & RAL & 51 & 58k (R) & RealSense cameras & $\checkmark$ (R) & External system for auto collection
            \\
            \hline
            TRANS-AFF\cite{jiang2022a4t} & 2022 & RAL & 8 & 1,346 (R) &RealSense cameras & $\times$ (R) & Affordance labelling
            \\
            \hline
            Evo-NeRF\cite{kerr2022evo} & 2022 & CoRL & 7 & 8,667 (S) & Blender, ZED Mini & $\checkmark$ (S), $\checkmark$ (R) & Rendered transparent objects with robust grasps
            \\
            \hline
            DREDS\cite{dai2022domain} & 2022 & ECCV  &1,861 & 130k (S) & Blender & $\checkmark$ (S) & Include raw depth in simulation
            \\
            \hline
            STD\cite{dai2022domain} &  2022 & ECCV  & 50 & 27k (R) &  RealSense D415 & $\checkmark$ (R) & Collection without external service
            \\ 
            \hline
        \end{tabular}}
                \begin{tablenotes}
        \small
        \item Note: S and R in \#Obj. and \#Imgs represent synthetic and real-world, respectively.
        \end{tablenotes}
\end{table*}

\section{Transparent Object Reconstruction}  \label{4}
Transparent object reconstruction methods that either reconstruct the noisy depth obtained with RGB-D cameras or reconstruct the 3D complete shapes of transparent objects can significantly mitigate the geometry gap between transparent objects and opaque objects, and further promote the outreach of robot application scenarios. For example, by using the reconstructed depth or 3D shapes, the grasping methods originally designed for opaque objects can be also applied to transparent objects.   

In this section, we first review the datasets published since 2020 for transparent object reconstruction. Then the state-of-the-art transparent object reconstruction methods are summarised. Finally, we highlight the main challenges of current transparent object reconstruction methods.  

\subsection{Datasets}
Transparent object reconstruction requires the ground truth of the reconstructed depth or 3D shape for evaluation. Therefore, datasets with ground truth of depth or 3D shapes are required for model training and evaluation.
In the subsection, we thoroughly summarise datasets published since 2020 for transparent object reconstruction, regarding year, place of publication (Pub.), number of objects in the images (\#Obj.), dataset size
(\#Imgs), devices, auto-collection ability and special features.

\noindent \textbf{ClearGrasp}~\cite{sajjan2020clear} dataset includes both a highly realistic synthetic dataset and a real-world benchmark. The synthetic dataset is rendered by using the ray-tracing Cycles rendering engine integrated into Blender, which can provide important effects for transparent objects, such as refraction and soft shadow.
To capture the depth of transparent objects in the real world, transparent objects are sprayed with rough stone textures that can reflect light evenly and lead to better depth estimates from RGB-D cameras.  
It should be noted that ClearGrasp is the first large-scale dataset including 50k synthetic images and 286 real images for the depth reconstruction of transparent objects.

\noindent \textbf{OOD}~\cite{zhu2021rgb} dataset consists of 60k synthetic images of five transparent objects from ClearGrasp~\cite{sajjan2020clear}. The Omniverse Platform and NVIDIA PhysX engine are used for rendering those images and getting natural poses of objects. To enhance the variety of the dataset, they augment the data by
changing textures for the ground, lighting conditions and camera views. 

\noindent \textbf{TODD}~\cite{xu2021seeing} dataset is a real-world depth reconstruction dataset that was created with an automated dataset creation workflow. First, a robotic arm equipped with an Intel RealSense RGB-D camera is controlled to multiple positions around the transparent objects for image collection. Then an automatic annotation system based on AprilTags~\cite{olson2011apriltag} is used to generate the ground truth of depth information.  
In total, TODD has 14,659 images of scenes collected with six similar glass beakers and flasks in five different backgrounds.

\noindent \textbf{TRANS-AFF}~\cite{jiang2022a4t} is a real-world transparent object dataset that is designed for both depth reconstruction and affordance detection. Similar to~\cite{sajjan2020clear}, the transparent objects in the original image are replaced with an identical spray-painted instance that can reflect light evenly to provide accurate depth information. In total, there are 1,346 pairs of RGB and depth images for 8 graspable containers.

\noindent \textbf{Dex-NeRF}~\cite{ichnowski2021dex} consists of 8 synthetic scenes generated with Blender Cycles and 8 real-world scenes captured with a Cannon EOS 60D camera and a RealSense camera. There are 5 and 6 different objects used for synthetic and real-world scenes, respectively. 
Every scene contains a set of images captured in a variety of camera poses.

\noindent \textbf{Evo-NeRF}~\cite{kerr2022evo} is a synthetic dataset of 8,667 rendered scenes of transparent objects in Blender simulation.  In total, 7 object meshes of common household transparent objects are used in the simulation.

\noindent \textbf{TransCG}~\cite{fang2022transcg} is a large-scale real-world dataset for transparent object depth reconstruction, which contains 57,715 RGB-D images from 130 different scenes. Similar to \textbf{TODD}, the images are collected with a robot equipped with cameras, i.e., RealSense D415 and RealSense L515. Moreover, a PST optical tracker system is used to estimate objects' poses and generate depth information.

\noindent \textbf{DREDS} and \textbf{STD} are the synthetic and real-world datasets proposed in~\cite{dai2022domain}. Specifically, DREDS is a synthetic dataset that consists of 130k domain randomised images of 1,861 different objects. Different from other synthetic datasets~\cite{sajjan2020clear, zhu2021rgb}, DREDS developed a simulated IR stereo camera to generate the raw depth scan that is noisy but similar to the one captured with physical depth cameras. STD is a real-world dataset that was captured with RealSense D415 and includes 27k images of 50 different objects. To generate the ground truth of the object's depth and pose, a photogrammetry-based reconstruction tool: Object Capture API\footnote{https://developer.apple.com/augmented-reality/object-capture/} was used.

\begin{remarknn}
In this subsection, 9 datasets for robotic transparent object reconstruction are reviewed and summarised in Table~\ref{tab:recon-dataset}. All datasets are either created through simulations using Blender or Omniverse, or gathered using physical sensors like RealSense D415 and ZED Mini. Blender is the most popular simulator, while RealSense cameras are the most frequently utilised physical sensors. In the following, we will discuss the aforementioned datasets in terms of their size, complexity, types of tasks and their special features.

\noindent \textbf{Dataset size and complexity.} 
Similar to the datasets in Sec.~\ref{3}, synthetic datasets are generally of a larger scale than real-world datasets. But differently, due to the application of automatic collection technology such as AprilTag~\cite{olson2011apriltag} and external localisation system, some real-world datasets such as TODD~\cite{xu2021seeing}, TransCG~\cite{fang2022transcg} and DREDS~\cite{dai2022domain} can also achieve a decent level of image quantity. According to the number of objects and images, we can observe that DREDS~\cite{dai2022domain} and TransCG~\cite{fang2022transcg} are the current most challenging synthetic dataset and real-world dataset, respectively.  

\noindent \textbf{Task types.} Except for the Dex-NeRF dataset~\cite{ichnowski2021dex}, all datasets in this subsection contain ground truth for depth maps and can be used as the benchmark for transparent object depth reconstruction. 
However, only several of them, i.e., Evo-NeRF dataset~\cite{kerr2022evo}, Dex-NeRF dataset~\cite{ichnowski2021dex}, and STD dataset~\cite{dai2022domain} can be used in 3D shape reconstruction such as visual hull based method and NeRF based methods, where multiple views of a single scene need to be kept in one category with known camera poses. 

\noindent \textbf{Special features.} These datasets also have their own characteristics as summarised in Table~\ref{tab:recon-dataset}. For example, TRANS-AFF~\cite{jiang2022a4t} includes affordance labelling that represents the object functionality of each pixel, which aids in the manipulation of transparent objects. Moreover, DREDS~\cite{dai2022domain} is the only dataset providing raw depth data in simulation, enabling the seamless sim-to-real transfer in the context of end-to-end depth reconstruction.

\end{remarknn}

\subsection{Approaches} 
\begin{figure*}[t]
  \includegraphics[width=\linewidth]{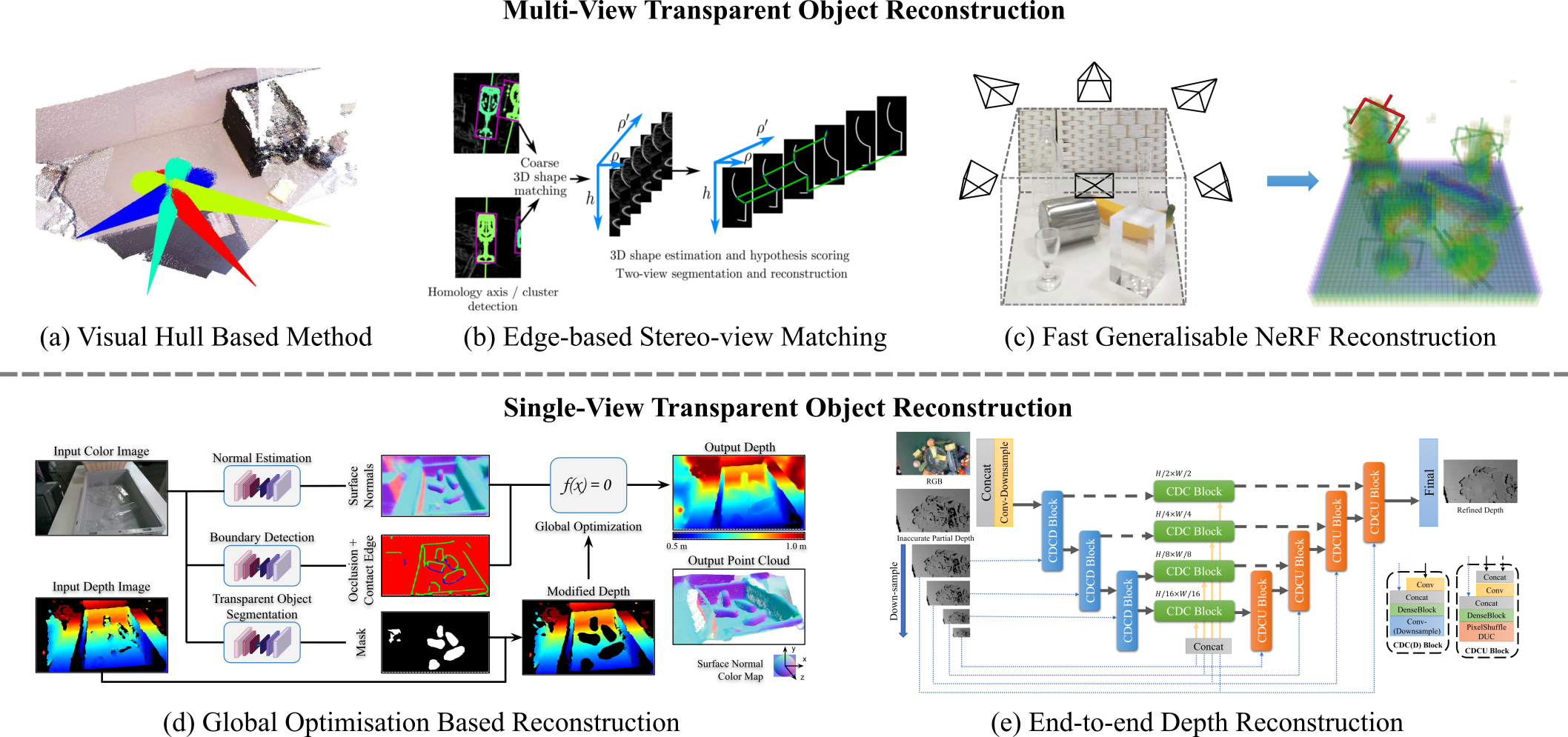}
  \caption{Different kinds of transparent object reconstruction approaches. \textbf{(a)} Voxel hull based method~\cite{albrecht2013seeing}; \textbf{(b)} Edge-based stereo matching~\cite{phillips2016seeing} (©[2016] MIT Press); \textbf{(c)} NeRF based Reconstruction~\cite{dai2022graspnerf} (©[2023] IEEE); \textbf{(d)} Global optimisation based reconstruction from ClearGrasp~\cite{sajjan2020clear} (©[2020] IEEE); \textbf{(e)} End-to-end Depth Reconstruction~\cite{fang2022transcg} (©[2022] IEEE).}
\label{fig:reconstruction}
\end{figure*}

Transparent object reconstruction approaches can be grouped into two categories based on the number of viewpoints used to capture the object: (1) multi-view approaches; (2) single-view approaches, as shown in Fig.~\ref{fig:reconstruction}. 
In this subsection, we will review both multi-view and single-view reconstruction methods that can be applied in robotic scenarios, and make a deep analysis from the perspectives of methodology. 

\subsubsection{Multi-view approaches}
Considering the limited information from a single view, most early studies focus on how to utilise the information gathered from multiple viewpoints in order to give an estimate of transparent object surfaces. 

\noindent \textbf{Visual hull based methods.} Inspired by the visual hull methods~\cite{de1999roxels, potmesil1987generating} that approximates an object's 3D shape by intersecting 3D projections of silhouettes from multiple 2D images, there are several similar methods proposed for transparent object reconstruction. In~\cite{albrecht2013seeing}, a frustum is first constructed for each individual frame based on the detected object shadow, and then the intersected part of all frusta is used to represent the transparent object shape, as shown in Fig.~\ref{fig:reconstruction}-(a). 
Similar to~\cite{albrecht2013seeing}, Torres and Mayol \cite{torres2014recognition} used intensity values to represent the possibility of the 3D point occupied by the glass, then used a threshold method to find the occupied space and reconstruct the geometry of the transparent object. In~\cite{ji2017fusing}, a silhouette-based visual hull reconstruction method was proposed to recover the lost surface of transparent objects. However, visual hull reconstruction from limited views might be inaccurate, besides missed concavities. In~\cite{li2020through}, the visual hull initialisation is refined based on the predictions of normals, total reflection mask and rendering error. It should be noted that this method may not be suitable for robotic applications as it requires the background map to be known prior and takes around 46 seconds to reconstruct a transparent shape from 10 views on an NVIDIA GeForce RTX 2080 Ti GPU.
To avoid the influence of reconstruction performance by the effect of motion blur on segmentation, a keyframe selection method was proposed in~\cite{zhu2021transfusion} to select the frames that contain very little or no motion blur as the input set. 

\noindent \textbf{Stereo-view matching methods.}
In~\cite{klank2011transparent}, a two-view reconstruction method was proposed that matches perspectively invariant features, and then triangulates the incorrect points in the stereo setup. 
In~\cite{phillips2016seeing}, rotationally symmetric transparent objects were reconstructed from two calibrated views with a set of contour points and tangents, as shown in Fig.~\ref{fig:reconstruction}-(b).

\noindent \textbf{NeRF based methods.} In the past two years, Neural Radiance Field (NeRF)~\cite{mildenhall2020nerf} that represents a 3D scene as a continuous function has been widely used to reconstruct the scene and synthesise novel view~\cite{pumarola2021d, hedman2021baking}. There are also several works about how to use NeRF to extract the geometry of challenging transparent objects. In~\cite{ichnowski2021dex}, NeRF encodes scene geometry using an MLP trained with 49 RGB images from different viewpoints, which is the first work specially designed to reconstruct the scene including transparent objects. 
However, it requires hours of computation for each scene, which deviates far from the real-time requirement of robotic applications. To address this time-consuming issue, Evo-NeRF~\cite{kerr2022evo} was proposed that takes advantage of the training speed of Instant-NGP~\cite{10.1145/3528223.3530127} and develops an active sensing approach to efficiently early stopping capture. Concurrently, Dai et al. in~\cite{dai2022graspnerf} leveraged one generalisable NeRF i.e., NeuRay~\cite{liu2022neural}, to achieve zero-shot construct NeRFs for novel scenes without training, as shown in Fig.~\ref{fig:reconstruction}-(c).

\begin{table*}
	\centering
		\caption{Comparison of the methods for transparent object reconstruction}
		\label{tab:reconstruction}
        \scalebox{1}{
		\begin{tabular}{ | p{0.10\linewidth} | p{0.12\linewidth} | p{0.2\linewidth} | p{0.2\linewidth} | p{0.25\linewidth}|        
         }
			\hline
		 Views &  Reconstruction Methods &  Advantages & Disadvantages & Examples \\
            \hline
        \multirow{3}{*}{Multiple View} & Visual Hull & Stable performance in indoor environments, straightforward implementation.  & Sensitive to calibration errors, need a good number of views, hard to cope with the concavity in object surface. & Albrecht et al.\cite{albrecht2013seeing}, Torres et al \cite{torres2014recognition}, Ji et al.\cite{ji2017fusing}, Transfusion~\cite{zhu2021transfusion}. \\
        \cline{2-5}
        & Stereo-view Matching & Require fewer views and lower computational cost. & Require external assumptions, e.g., the object's symmetry and planarity. & Klank et al.~\cite{klank2011transparent}, Seeing Glassware\cite{phillips2016seeing}. \\
        \cline{2-5}
        & NeRF & Robust to transparent objects with different shapes, like the glass containers with concavity.  & Heavy computational cost, need a good number of views, sensitive to calibration errors. & Dex-NeRF~\cite{ichnowski2021dex}, Evo-NeRF~\cite{kerr2022evo}, GraspNeRF~\cite{dai2022graspnerf}.\\
        \hline
        \multirow{2}{*}{Single View} & Global Optimisation & Generalised well to unseen objects, and require only a single RGB-D image. & Sensitive to lighting changes, easily influenced by occlusion.& ClearGrasp~\cite{sajjan2020clear}, A4T~\cite{jiang2022a4t}.\\
        \cline{2-5}
        & End-to-end ~~~~~~~Reconstruction & Fast reconstruction speed, good performance in structured environments. & Require large-scale datasets, and not robust to cluttered environments especially when two objects are overlapped. & RGB-D Implicit~\cite{zhu2021rgb}, TranspareNet~\cite{xu2021seeing}, DepthGrasp~\cite{tang2021depthgrasp}, DFNet~\cite{fang2022transcg}, SwinDRNet~\cite{dai2022domain}, TODE-Tran~\cite{chen2022tode}.\\
        \hline
                     
        \end{tabular}}
\end{table*}

\subsubsection{Single-view approaches}
While some of the previously mentioned methods can achieve fast reconstruction by introducing a set of tricks, the long time caused by capturing images from multiple views cannot be avoided. Hence, some researchers focus on how to achieve the fast and accurate reconstruction of transparent objects with the image captured from a single view. Many single-view approaches often require specialised capturing systems~\cite{miyazaki2004transparent, rantoson20103d,  yeung2014normal, han2015fixed} or known background patterns~\cite{qian20163d, wu2018full} to achieve a good reconstruction of transparent objects~\cite{ihrke2010transparent}. However, those strong assumptions make them hard to apply to scenarios where robots work, such as transparent object manipulation. There are also several approaches to reconstructing transparent objects not being constrained by the aforementioned assumptions, e.g., optimisation based reconstruction methods and end-to-end reconstruction methods, as introduced below.

\noindent \textbf{Optimisation based methods.} Sajjan et al.~\cite{sajjan2020clear} used the global optimisation algorithm proposed in~\cite{zhang2018deep} to reconstruct the missing or noisy depth regions of transparent objects based on the predicted surface normals, as shown in Fig.~\ref{fig:reconstruction}-(d). However, when the reconstructed regions are enclosed by occlusion boundaries, the reconstructed depths become indeterministic and can be assigned random values. To address this issue, an affordance-based multi-step depth reconstruction method was proposed in~\cite{jiang2022a4t} that combines both the global optimisation algorithm proposed in~\cite{zhang2018deep} and RANSAC-based plane fitting method. Nevertheless, the above two methods are constrained by the prediction of surface normals and contact edges. In cluttered environments, the occlusions between different objects may lead to noisy predictions of surface normals and invisible contact edges, which results in failed depth reconstruction.

\noindent \textbf{End-to-end methods.} To overcome the above drawback, there are several works utilising end-to-end reconstruction methods that are not dependent on surface normals and contact edges.
In~\cite{zhu2021rgb}, Zhu et al. used a local implicit neural representation and an iterative depth refinement model to complete the depth information of transparent objects. 
Xu et al. \cite{xu2021seeing} proposed a joint point cloud and depth completion method named TranspareNet to leverage RGB and depth signals of transparent objects.
In~\cite{tang2021depthgrasp}, a generative adversarial network named DepthGrasp was proposed that utilises the generator to reconstruct the depth maps of transparent objects. 
As shown in Fig.~\ref{fig:reconstruction}-(e), a multi-scale deep network was proposed in~\cite{fang2022transcg} that takes an RGB image concatenated with inaccurate partial depth as input. 
To extract fine-grained feature representation for depth reconstruction, transformer~\cite{vaswani2017attention} was utilised in TODE-Trans~\cite{chen2022tode} which is an encoder-decoder framework. 
In~\cite{dai2022domain}, a two-stream Swin Transformer~\cite{liu2021swin} based RGB-D fusion network, SwinDRNet, was proposed for learning to perform depth restoration. Different from TODE-Trans which predicts the depth map directly, SwinDRNet fuses the raw depth and predicted depth using the predicted confidence map. 

\begin{remarknn} Table~\ref{tab:reconstruction} summarises the comparison of reconstruction methods for transparent objects. Single-view methods are generally more efficient than multi-view approaches due to their elimination of camera movement and lower computational costs. However, single-view techniques rely heavily on high-quality datasets and can suffer from environmental factors such as poor lighting conditions and cluttered objects, which limit their robustness compared to multi-view methods. Future research can explore strategies for enhancing the robustness of single-view methods to these variations.  
\end{remarknn}

\subsection{Challenges and Open Questions}
In this subsection, we discuss three key challenges for transparent object reconstruction, i.e., the difficulties of getting the ground truth, the design of reconstruction algorithms, and reconstruction in challenging environments.
\subsubsection{Getting the ground truth in the real world}
Unlike the ground-truth masks for transparent object segmentation, the ground truth for reconstruction is hard to be annotated by human annotators. 
Although various approaches have been adopted to generate the ground truth of depth maps, e.g., replacing transparent objects with opaque counterparts sharing identical shapes and positions~\cite{sajjan2020clear, jiang2022a4t}, utilising AprilTag arrays~\cite{xu2021seeing} or an external optical tracking system~\cite{fang2022transcg} for localisation, and proposing an interactive GUI for annotators to label object poses~\cite{chen2022clearpose}, each approach presents its own set of challenges. These issues include time-consuming processes~\cite{sajjan2020clear,chen2022clearpose} and the need for external markers~\cite{xu2021seeing, fang2022transcg}, making them less than ideal solutions.

To address the labelling issue for real-world datasets, one possible solution is utilising other advantaged sensors. It has been proved that large-scale tactile sensors are capable of assisting the visual perception system for human pose estimation~\cite{luo2021intelligent}, which makes it possible to label the transparent object pose without externally attached markers as in~\cite{xu2021seeing, fang2022transcg}. Another possible solution is utilising self-supervised learning~\cite{zhu2017unpaired, 9594655, 9229197} to avoid the requirement of paired depth images of transparent and opaque objects. 

\begin{table*}
\centering
 \begin{threeparttable}
	
		\caption{Comparison of the datasets for transparent objects pose estimation}
		\label{tab:otherdataset}
        
		\begin{tabular}{ | p{0.14\linewidth} | p{0.03\linewidth} | p{0.05\linewidth} | p{0.05\linewidth} | p{0.07\linewidth}| p{0.10\linewidth}| p{0.12\linewidth}| p{0.11\linewidth} |p{0.09\linewidth} |}
			\hline
			Dataset & Years & Pub. & \#Obj. & \#Imgs & Modalities & Devices & Scene Type & Outdoor Environments\\
            \hline
            ProLIT~\cite{zhou2020lit} & 2020 & RAL & 442 & 75k (S), 300 (R)& Light-field images & Unreal Engine, Lytro Camera   & Single object & $\times$
            \\
            \hline
            TOD~\cite{liu2020keypose} & 2020 & CVPR & 15 & 28k (R) & Stereo, RGB-D & RealSense D415, ZED camera & Single object &$\times$\\
            \hline
            StereOBJ-1M\cite{liu2021stereobj} & 2021 & ICCV & 7 & 393k (R) & Stereo & Weeview camera & Cluttered object  
            & $\checkmark$\\
            \hline
            ClearPose\cite{chen2022clearpose} & 2022 & ECCV & 63 & 350k (R) & RGB-D & RealSense L515 & Cluttered object &  $\times$
            \\
            \hline
            PhoCaL\cite{wang2022phocal} & 2022 &  CVPR & 8 & 3,951 (R) & Polarisation, RGB-D & RealSense L515, Lucid Phoenix &Cluttered object
            & $\times$\\
            \hline
            Syn-TODD\cite{wang2023mvtrans} & 2023 & ICRA & 16k
 & 113k(S) & RGB-D & Blender & Cluttered object & $\times$ \\
            \hline
        \end{tabular}
        \begin{tablenotes}
        \small
        \item Note: only transparent objects are included in the object count.  S and R in \#Obj. and \#Imgs represent synthetic and real-world, respectively.
        \end{tablenotes}
        \end{threeparttable} 
\end{table*}

\subsubsection{Design of reconstruction algorithms}
As outlined in Table~\ref{tab:reconstruction}, NeRF-based reconstruction methods and single-view end-to-end reconstruction methods face challenges with respect to efficiency and robustness, respectively. Hence, three potential research directions could be: (1) accelerating NeRF-based reconstruction methods speed by exploring techniques such as learning-based sampling, and efficient network architectures; (2) enhancing the generalisation and robustness of single-view end-to-end methods by integrating attention mechanisms or other context-aware components into the model architecture; (3) developing novel methods that leverage the efficiency of single-view techniques while incorporating the robustness of multi-view approaches could result in more accurate and reliable transparent object reconstruction models, suitable for a wider range of applications.

Moreover, as discussed in Section~\ref{3}, synthetic data often lack realistic feature artefacts (e.g., sensor noise, true lighting etc.), which introduces the domain gap between simulations and the real world. To overcome this Sim2Real gap, domain randomisation methods have been thoroughly investigated in several works~\cite{tobin2017domain, sajjan2020clear, jiang2022robotic, eppel2022predicting}. However, it is still questionable whether current domain adaptation approaches~\cite{zhao2019geometry, 9851499} can improve the reconstruction performance further.  

\subsubsection{Reconstruction in challenging environments} 
Current reconstruction methods show good performance in structured environments where lighting conditions are sufficient and objects are not occluded by each other. However, this assumption is hard to be met in real robotic applications. For example, the glass cups to be sorted in the washing machine are randomly placed and may occlude with each other. Therefore, it is still a challenging problem to reconstruct transparent objects in challenging environments. 

\noindent \textbf{Environments with extreme lighting.}
Vision-based reconstruction methods are susceptible to changes in lighting conditions. As discussed in~\cite{sajjan2020clear}, bright directional lighting and its associated caustics cause the mistaken prediction of surface normals and segmentation masks, eventually leading to a failure of reconstruction of transparent objects. 
To overcome the challenges of extreme lighting conditions, designing light-invariant features and using other modalities that are insusceptible to light changes could be alternative solutions. 

\noindent \textbf{Cluttered environments.}
Highly cluttered environments where multiple objects are partially or completely occluding each other were recognised as the main challenges in different robotic applications, such as Amazon Picking Challenge~\cite{correll2016analysis}, and UAV path planning~\cite{oleynikova2018safe}. It becomes even more challenging when the cluttered objects are transparent, as some image pixels could belong to more than one object because of transparency. It is still an open problem to reconstruct transparent objects in highly cluttered environments.

\section{Transparent Object Pose Estimation}\label{5}
Transparent object pose estimation refers to the task of determining the position and orientation of a transparent object in 3D space. It is crucial for various applications, including augmented reality, robotics, and computer graphics, where the accurate positioning of transparent objects is essential. In this section, we first review the recent datasets for transparent object pose estimation. Then transparent object reconstruction methods are summarised. Finally, we summarise the challenges and open questions for transparent object perception. 

\subsection{Dataset}
In this subsection, we thoroughly summarise recent datasets for pose estimation. Since some datasets are designed for both depth reconstruction and pose estimation such as ClearGrasp~\cite{sajjan2020clear} and TransCG~\cite{fang2022transcg}, the datasets that have already been introduced in the previous sections will not be repeated here.

\noindent \textbf{ProLIT}~\cite{zhou2020lit} is a light-field image dataset mainly for the task of transparent object 6D pose estimation. This dataset contains a total of 75,000 synthetic images generated with Unreal Engine and 300 real-world images captured with a Lytro Illum camera. This dataset is labelled with pixel-wise semantic segmentation and 6D object poses. 

\noindent \textbf{TOD}~\cite{liu2020keypose} is
a real-world dataset of 15 clear objects in five classes, with 48k 3D-keypoint labelled images for transparent object pose estimation. To automatically capture the images from different views, a robotic arm equipped with both a Kinect Azure camera and a Stereolabs ZED camera is used.
However, TOD records in a studio environment and does not include occluded objects.

\noindent \textbf{StereOBJ-1M}~\cite{liu2021stereobj} is the largest 6D object pose dataset that consists of 396,509 high-resolution stereo frames and over 1.5 million 6D pose annotations of 18 objects recorded in 183 indoor and outdoor scenes. Compared to TOD, StereOBJ-1M allows the objects to be placed in more flexible and complex background terrains. However, it only includes 7 transparent objects in the same category, i.e., plastic rack.

\noindent \textbf{ClearPose}~\cite{chen2022clearpose} is a large-scale dataset for segmentation, scene-level depth completion and object-centric pose estimation tasks. ClearPose dataset contains over 350k labelled real-world RGB-Depth frames and 5M instance annotations covering 63 household objects. Different from~\cite{xu2021seeing} that estimates the pose of transparent objects with AprilTag, an add-on in Blender~\cite{chen2022progresslabeller} was used to realise rapid data annotation and exempts from the broken depth problem by transparent objects.

\noindent \textbf{PhoCaL}~\cite{wang2022phocal} is a multi-modal dataset for category-level object pose estimation of photometrically challenging objects including transparent objects. 
It captured 3,951 frames of RGB-D images and polarised images for 8 transparent objects. 
With the manipulator-driven annotation pipeline, PhoCaL reaches pose accuracy levels that are one order of magnitude more precise than previous vision-sensor-only pipelines even for photometrically complex objects.

\noindent \textbf{Syn-TODD}~\cite{wang2023mvtrans} is a dataset comprising 113,772 stereo image pairs of 1,996 distinct scenes that feature a combination of 9,012 unique opaque objects, along with 7,010 unique transparent objects generated procedurally. Syn-TODD has broad compatibility with various methods such as RGB, RGB-D, stereo, and multi-view based pose estimation techniques.

\begin{remarknn}
In this subsection, six datasets for transparent object pose estimation are reviewed and summarised in Table \ref{tab:otherdataset}. In the following, We discuss the aforementioned datasets in terms of their types of tasks, and their overall qualities.

\noindent \textbf{Dataset tasks.}
StereOBJ-1M~\cite{liu2021stereobj} and PhoCal~\cite{wang2022phocal} datasets are not specifically designed for transparent object pose estimation and instead include a wide range of objects, such as transparent, specular, and opaque objects. Hence, both of them can be used as benchmarks for general object pose estimation. In contrast, ProLIT~\cite{zhou2020lit}, TOD~\cite{liu2020keypose}, and ClearPose~\cite{chen2022clearpose} including only transparent objects are more appropriate as benchmarks for transparent object pose estimation. It should be also noted that the datasets such as ClearGrasp~\cite{sajjan2020clear} and TransCG~\cite{fang2022transcg} intended for depth reconstruction also contain object pose annotations, and could be used for pose estimation. 

\noindent \textbf{Dataset quality.}
ClearPose~\cite{chen2022clearpose} and Syn-TODD~\cite{wang2023mvtrans} have been recognised as the most challenging real-world and synthetic datasets for transparent object pose estimation due to their large number of objects and images, as well as the diverse and complex scenes they include. However, the main drawback is that both of them only include RGB-D images, which may limit the types of models that can be trained and tested on the data. It should be also noted that ProLIT~\cite{zhou2020lit} is the only dataset that includes both synthetic and real-world data, which makes it a unique and valuable resource for evaluating transparent object pose estimation algorithms.   
\end{remarknn}

\begin{table*}
	\centering
		\caption{Comparison of the methods for transparent object pose estimation}
		\label{tab:poseestimation}
        \scalebox{1}{
		\begin{tabular}{ | p{0.10\linewidth} | p{0.11\linewidth}|p{0.20\linewidth} | p{0.25\linewidth} | p{0.20\linewidth} |  
  }
			\hline
			Solver Type &  Methods & Advantages & Disadvantages & Examples \\
            \hline
             &  Silhouette-based matching  & Easy implementation, robust to light changes. & Long-time sampling, requiring a known object's model. & Lysenkov et al.\cite{lysenkov2013pose, lysenkov2013recognition}, LIT~\cite{zhou2020lit} \\
            \cline{2-5}
            & Boundary-based matching & Low computational cost.  & Require the assumption of object shape. & Seeing Glassware\cite{phillips2016seeing}\\
            \cline{2-5}
            \multirow{2}{*}{Indirect solver} & Keypoint-based pose estimation & Light-scale representation, fast-speed pose estimation. & Sensitive to occlusions. &  KeyPose\cite{liu2020keypose}, GhostGrasp\cite{chang2021ghostpose}, Byambaa et al.\cite{byambaa20226d}.
            \\
            \cline{2-5}
             & NOCS-based PnP & Robust to occlusions, and applicable to category-level pose estimation.& Heavy computational loads, require large-scale datasets to train the model. & StereoPose~\cite{chen2022stereopose}
            \\
            \hline 
            Direct solver & End-to-end ~~~~~~regression & Applicable to category-level pose estimation. & Rely on the depth reconstruction accuracy, or additional modality, i.e., polarisation for NOCS prediction. & Xu et al.\cite{xu20206dof}, TransNet~\cite{zhang2022transnet}, PPP-Net~\cite{gao2022polarimetric}, MVTrans~\cite{wang2023mvtrans}.\\
            
            \hline

        \end{tabular}}
\end{table*}

\subsection{Approaches}
Object pose estimation is the crux of many important real-world applications, such as robotic grasping and manipulation. 
As summarised in Table~\ref{tab:poseestimation}, current leading approaches for transparent object pose estimation either directly regress object pose from images~\cite{xu20206dof,zhang2022transnet} or indirectly solve the pose via 3D model matching~\cite{lysenkov2013recognition, lysenkov2013pose, zhou2020lit}, triangulation methods~\cite{chang2021ghostpose} or Perspective n Points (PnP) ~\cite{byambaa20226d, chen2022stereopose}. 

\subsubsection{Indirect methods} The early studies on transparent object pose estimation mainly focused on indirect methods that do not directly compute the object's pose but instead rely on intermediate representations or cues such as silhouette, contour, and keypoints. 
In~\cite{lysenkov2013recognition}, a silhouette-based matching method was proposed to generate an initial pose which is then refined with the support plane assumption and the fitting of a 3D model.
To estimate the pose of cluttered objects, Lysenkov et al. in~\cite{lysenkov2013pose} used Geometric Hashing and an edge-based 3D model fitting to replace the silhouette segmentation method and the refine method in~\cite{lysenkov2013recognition}, respectively. 
Similar to~\cite{lysenkov2013pose} that used edge-based fitting method, \cite{phillips2016seeing} localise the rotationally symmetric object by matching the detected edges from two calibrated views.

Instead of using the common visual features such as edge and silhouette, Zhou et. al. in~\cite{zhou2018plenoptic} propose a new descriptor, Depth Likelihood Volume (DLV), to address the uncertainties from the translucency by generating possible depth likelihoods for each pixel. Then, the six-DoF object pose is estimated by Monte Carlo Localisation over a constructed DLV. To reduce the computational load, a two-stage method was proposed in~\cite{zhou2020lit} that leverages the power of discriminative and generative methods and calculates the 6D pose of transparent objects in a sampling-based iterative likelihood re-weighting process. 
Moreover, Liu et. al. in~\cite{liu2020keypose} used keypoints to represent the transparent object pose instead of calculating the 6-DOF pose. In~\cite{liu2020keypose}, KeyPose was proposed to predict 3D keypoints on transparent objects from cropped stereo RGB input utilising an early fusion technique. Following the KeyPose work~\cite{liu2020keypose}, Chang et. al. in~\cite{chang2021ghostpose} used the generalised keypoints extracted from the predicted 3D bounding box and multi-view information to estimate the 3D keypoints of transparent objects. 

There are also several studies that used PnP pipeline to optimise the displacement between 2D-3D correspondences, i.e., the correspondence between image and 3D model. In~\cite{byambaa20226d}, the 2D keypoints are estimated using a deep neural network. Then, the PnP algorithm takes camera intrinsics, object model size, and keypoints as inputs to estimate the 6D pose of the object. However, the keypoint based methods may suffer from noise when the object shape and size are various. Hence, dense 2D-3D correspondence i.e., Normalised Object Coordinate Space (NOCS), has been used as a representation for transparent object pose estimation~\cite{chen2022stereopose}. In~\cite{chen2022stereopose}, the NOCS maps from both the front view and back view are utilised in a PnP algorithm to predict the pose of transparent objects. The back-view NOCS maps can significantly address the problem of image content aliasing for transparent objects. 

\subsubsection{Direct methods}
Beyond those indirect ways mentioned above, there are also several studies directly regressing object pose. Xu et. al.~\cite{xu20206dof} fed the accurate point cloud reconstructed with ClearGrasp~\cite{sajjan2020clear} to a DenseFusion-like network~\cite{wang2019densefusion} for predicting the 6-DOF pose of a transparent object. Zhang et. al. in~\cite{zhang2022transnet} fed the point cloud reconstructed with DFNet~\cite{fang2022transcg} to PointFormer~\cite{zou20226d} for predicting both the pose and scale of transparent objects, simultaneously.  
Thanks to the rapid development of AI, differentiable pose estimators have been proven to achieve higher accuracy than RANSAC/PnP~\cite{wang2021gdr, chen2022epro}. Hence, in~\cite{gao2022polarimetric}, both the NOCS maps and surface normals predicted with a hybrid model (vision and polarisation) were used to regress the transparent object pose. Zhang et al.~\cite{wang2023mvtrans} proposed an end-to-end multi-view architecture named MVTrans with multiple perception capabilities, including depth estimation, segmentation, and pose estimation.

\begin{remarknn}
We compare different methods for transparent object pose estimation in Table~\ref{tab:poseestimation}. It has been observed that earlier approaches, such as silhouette-based and boundary-based matching methods, have experienced a decline in popularity due to their reliance on prior information like object models or symmetry. In recent years, keypoint-based pose estimation methods and end-to-end pose estimation methods have gained significant interest in the field of transparent object pose estimation with excellent performance. However, these methods have their limitations: keypoint-based pose estimation methods are susceptible to occlusions, whereas the performance of end-to-end pose estimation methods relies on the accuracy of depth reconstruction or NOCS estimation.   
\end{remarknn}

\begin{figure*}[t]
  \includegraphics[width=\linewidth]{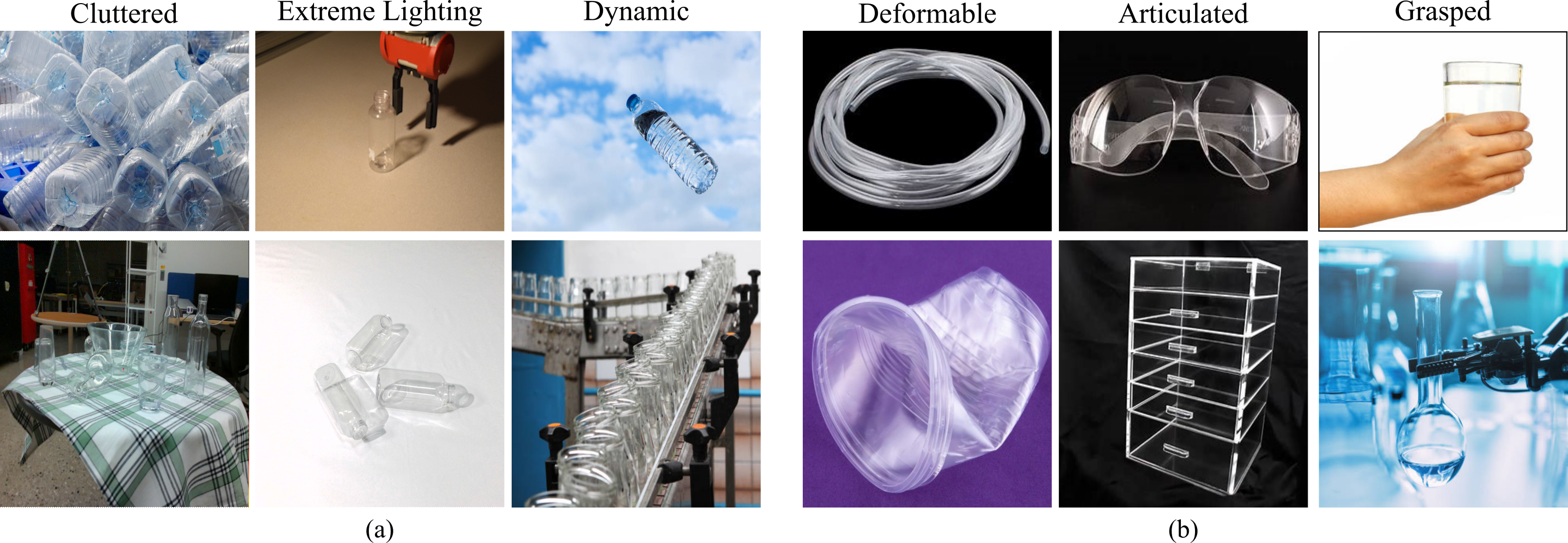}
  \caption{Different kinds of challenges summarised in this article. \textbf{(a)} Environment-level challenges that include cluttered environment, extreme lighting environment, and dynamic environment; \textbf{(b)} Object-level challenges that include deformable objects, articulated objects, and grasped objects. }
\label{fig:challenges}
\end{figure*}

\begin{table*}
        \setlist[itemize]{wide=0pt, nosep, leftmargin= *, after=\vspace{-\baselineskip}}
	\centering
		\caption{An overview of the challenges and open questions.}
		\label{tab:challenges}
        \scalebox{1}{
		\begin{tabular}{  | p{0.11\linewidth}|p{0.30\linewidth} | p{0.50\linewidth} |}
			\hline
		  Topics &  Challenges & Open Questions \\
            \hline
             Dataset ~~~~~~~~preparation  & Limited scales of real-world transparent object datasets. & \begin{itemize} \item Simulating multiple modalities of data e.g., polarised images and thermal images. \item Learning domain-invariant features for Sim2Real transfer. \item Designing efficient labelling tools for transparent object segmentation. \item Using few-shot learning to train a robust model with a small amount of data. \item Using weakly supervised learning to train a model with partially or weakly labelled data.  \end{itemize}\\
             \cline{2-3}
             & Difficulties of getting the ground truth for depth reconstruction. & \begin{itemize} \item Using external sensors to obtain the ground truth. \item Utilising self-supervised learning to avoid the requirement of paired depth images of transparent objects and opaque objects. \end{itemize} \\
             \hline
            Challenging ~~~~~~~environments & Real-time requirements in dynamic environments. & \begin{itemize}
                \item Using the sensing modalities that can reflect the physical properties of transparent objects.
                \item Representing sensing modalities in an efficient way.
                \item Accelerating the inference speed with pruning and quantisation techniques. 
                \item Using the temporal information between different frames.
            \end{itemize} \\
            \cline{2-3}
             & Perception of transparent objects in extreme lighting conditions. & \begin{itemize} \item Designing light-invariant features. \item Using multi-expert learning to train a model that is robust in different lighting conditions. \item Utilising modalities that are insusceptible to light changes, such as tactile sensing and polarisation images.  \end{itemize} \\
            \cline{2-3}
             & Perception of transparent objects in a highly cluttered environment. & \begin{itemize} \item Fusing multiple modalities to enhance the robustness of reconstruction methods. 
             \item Active depth reconstruction of transparent objects.\end{itemize} 
             \\
             \hline
             Challenging ~~~objects &  Grasped transparent objects are severely occluded by the robot hand or the object itself. & \begin{itemize} \item Using auxiliary information such as the pose of the robot hand. \item Using tactile sensing to assist the visual perception. \item Perception from the interaction. \end{itemize} \\
            \cline{2-3}
             & Changeable shapes of deformable transparent objects or articulated transparent objects. & \begin{itemize} \item Designing an optimal pose or shape representation for the objects with changeable shapes. \item Using Recurrent Neural Network (RNN) where temporal information is involved to perceive such objects.  \end{itemize} \\
            \hline
        \end{tabular}}
\end{table*}

\subsection{Challenges and Open Questions}
Parts of the challenges for transparent object pose estimation and grasping overlapped with the aforementioned challenges in Section~\ref{3} and Section~\ref{4}, e.g., time-consuming dataset generation, and challenging environments. To this end, we will only discuss other challenges that have not been investigated for transparent objects in this section, i.e., the design of pose estimation methods for transparent objects, and perception of challenging objects.
\subsubsection{Design of pose estimation methods}
In recent years, NOCS-based methods with the ability to handle occlusions and partial views of objects have been at the forefront of research on the pose estimation of opaque objects. However, transparent object pose estimation tends to rely on keypoint representations~\cite{chang2021ghostpose}, reconstructed depth~\cite{xu20206dof} or end-to-end estimator~\cite{wang2023mvtrans}. A potential explanation for this circumstance is that transparent objects without distinct features can negatively influence NOCS prediction, ultimately resulting in poor pose estimation accuracy. Therefore, potential research directions for transparent object pose estimation could be: (1) utilising additional modalities to improve NOCS estimation of transparent objects; (2) designing robust pose estimators that can handle NOCS estimation errors; (3) developing optimised representations for transparent objects, such as combining keypoints with NOCS, to enhance the accuracy and robustness of pose estimation algorithms. 

\subsubsection{Perception of challenging objects} There are several challenges associated with the state and geometries of transparent objects. 

\noindent \textbf{Deformable or articulated transparent objects.} One challenge is the perception of transparent objects with changeable shapes such as deformable transparent objects (disposable plastic cups, and plastic tubings) and articulated transparent objects (acrylic drawers, glasses with transparent frames). As discussed in~\cite{zhu2022challenges}, deformable object perception is an emerging research problem in robotics. It would be very interesting to consider both the deformation and transparency when perceiving deformable transparent objects. 

\noindent \textbf{Grasped transparent objects.} Another challenge is the pose estimation of grasped transparent objects, which is essential for dexterous manipulation. For example, robot assistants need to be capable of monitoring the status of the grasped transparent cup and the relative pose between the cup and people's mouths when assisting people to drink. It is still questionable whether current pose estimation approaches for grasped opaque objects are applicable to transparent objects.   

\noindent \textbf{Transparent objects in dynamic environments.} The previous studies on transparent object pose estimation mainly focus on static environments. However, in the waste recycling factory, the plastic bottles to be sorted are moving on a conveyor belt. The dynamics pose additional challenges for transparent object pose estimation, i.e., moving blur, computation delay, and motion uncertainty.

\section{Summary of Challenges and Open Problems}
The challenges related to environments and objects discussed in Sec.~\ref{3},~\ref{4}, and \ref{5} can be summarised in Fig.~\ref{fig:challenges}. To facilitate the development of the robotic perception of transparent objects, the overview of the challenges and open questions are summarised in Table~\ref{tab:challenges}. 

Apart from the topics discussed earlier, researchers are also turning their attention to tasks such as transparent object tracking and grasping. Zhou et al.~\cite{8967685} used using plenoptic sensing to detect the grasp pose of transparent objects. Weng et al.~\cite{weng2020multi} have employed transfer learning to adapt grasping models trained on depth maps to transparent object grasping with RGB images. 
Jiang et al.~\cite{2022jiaqi} used the GelSight tactile sensor to help vision determine the grasp pose of transparent objects. 
However, the limited amount of research on these tasks hinders comprehensive analysis. Expanding the scope of research on transparent object tracking and grasping could yield a more profound understanding of their potential applications and impact. For example, current research on transparent object manipulation has not yet considered the challenges posed by the misalignment between the visual centroid and centre of mass (CoM)~\cite{veres2020incorporating, liang2023visuo}. Investigating the CoM of transparent objects could be a promising direction, potentially enhancing the efficiency and robustness of transparent object grasping.

In addition to developments of perception algorithms, we anticipate the emergence of novel sensors designed especially for transparent object perception, as well as the exploration of new modalities. For example, mm-wave radar is gaining popularity as an emerging technology for robotic applications, such as Simultaneous Localisation and Mapping~\cite{kramer2022coloradar} and human behaviour monitoring~\cite{sengupta2020mm}, and its potential for transparent object perception warrants further investigation.

\section{Conclusion}
In this survey, we provide a comprehensive review of the sensors and simulation software used in transparent object perception, as well as a detailed analysis of several major tasks of transparent object perception, including transparent object segmentation, reconstruction, and pose estimation. For each task, we have introduced various subdivided methods, related datasets and open research questions. 
Our interactive online website allows readers to easily navigate the datasets and methods presented in each reference, facilitating further exploration and research in the field of transparent object perception.


\bibliographystyle{IEEEtran}
\bibliography{egibib}

\vfill

\end{document}